\documentclass[final,3p,times,twocolumn]{elsarticle}

\usepackage{epsfig}

\usepackage{graphicx}%
\usepackage{multirow}%
\usepackage{amsmath,amssymb,amsfonts}%
\usepackage{amsthm}%
\usepackage{mathrsfs}%
\usepackage[title]{appendix}%
\usepackage{textcomp}%
\usepackage{manyfoot}%
\usepackage{booktabs}%
\usepackage{algorithmicx}%
\usepackage{algpseudocode}%
\usepackage{listings}%
\usepackage[ruled,vlined]{algorithm2e}
\usepackage{color, colortbl}
\usepackage{siunitx}

\usepackage{booktabs}
\usepackage{multirow}
\usepackage[table,xcdraw]{xcolor}
\usepackage{times,epsf,mathtools,xcolor,amssymb,amsmath,hyperref, enumitem,wrapfig, subcaption, comment, cleveref, appendix, float}
\usepackage{tabu}
\usepackage{xcolor}
\usepackage{graphicx}
\usepackage{tikz}
\usetikzlibrary{shapes.geometric, arrows}

\newcommand{\PD}{\mathrm{PD}}

\newcommand{\wh}{\widehat}

\newcommand{\X}{\mathcal{X}}

\newcommand{\Y}{\mathcal{V}}

\definecolor{LightCyan}{rgb}{0.88,1,1}

\begin{document}

\begin{frontmatter}


 \ead{ahmedf9@erau.edu}

\title{Topological Signatures vs. Gradient Histograms: A Comparative Study for Medical Image Classification}


\author[aff1]{Faisal Ahmed}

\address[aff1]{Department of Data Science and Mathematics, Embry-Riddle Aeronautical University, 3700 Willow Creek Rd, Prescott, Arizona 86301, USA}

\begin{abstract}
This study presents the first comparative evaluation of two fundamentally different feature extraction methods—\emph{Histogram of Oriented Gradients (HOG)} and \emph{Topological Data Analysis (TDA)}—for classifying medical images, with a focus on retinal fundus imagery. HOG captures local shape information by computing the distribution of gradient orientations within small spatial regions, effectively encoding texture and edge features. In contrast, TDA, specifically through \emph{cubical persistent homology}, extracts high-level topological descriptors that quantify the global shape and connectivity of pixel intensities across the image.

We apply these techniques to the large-scale, publicly available APTOS dataset to address two classification tasks: a binary classification (normal vs. diabetic retinopathy (DR)) and a five-class DR severity grading. From each fundus image, we extract \num{26244} HOG features and \num{800} TDA features, which are then used independently to train seven classical machine learning models (logistic regression, random forest, XGBoost, SVM, decision tree, $k$-NN, and Extra Trees) using 10-fold cross-validation.

\emph{XGBoost} consistently outperformed other models, achieving an accuracy of \textbf{94.29$\pm$1.18\%} with HOG features and \textbf{94.18$\pm$0.92\%} with TDA features for the binary task. In the multi-class setting, it obtained \textbf{74.41$\pm$1.02\%} (HOG) and \textbf{74.69$\pm$1.52\%} (TDA). These results demonstrate that while both methods provide competitive performance, they encode fundamentally different aspects of image structure.

This work is the first to systematically compare \emph{gradient-based} and \emph{topological feature extraction} for fundus image classification. Beyond their standalone utility, both feature sets can be integrated into deep learning pipelines to enhance diagnostic accuracy while maintaining interpretability. Our findings lay the groundwork for hybrid approaches that combine geometric and topological insights in medical image analysis. While our experiments were conducted on retinal images, the proposed methodology is generalizable to a wide range of medical imaging modalities.

\end{abstract}


\begin{highlights}
    \item We conduct the first systematic comparison of two handcrafted feature extraction techniques—\emph{persistent homology} from topological data analysis (TDA) and \emph{Histogram of Oriented Gradients} (HOG)—for medical image classification.
    
     \item We identify and address a gap in the current literature: while both HOG and TDA have been applied independently in medical imaging tasks, no prior study has directly compared their performance on the same dataset or disease classification problem. Our work provides the first such comparison.

    \item The evaluation is performed on the large-scale APTOS dataset under two classification tasks: (i) binary classification (normal vs.~diabetic retinopathy), and (ii) multi-class DR severity grading across five levels.
    
    \item Each fundus image is represented by 800 topological features and $26{,}244$ HOG features, which are independently assessed using seven classical machine learning models, including XGBoost, SVM, and random forest.
    
    \item Results highlight the complementary strengths of topological and gradient-based features, offering insights into their respective advantages for medical image analysis tasks.
    
    \item Both TDA and HOG are lightweight, interpretable, and readily integrable into deep learning pipelines, enabling future hybrid approaches that combine geometric and topological representations for improved diagnostic performance.
\end{highlights}


\end{frontmatter}



\section{Introduction}
\label{sec:introduction}

Medical imaging plays a crucial role in modern healthcare, enabling early detection, diagnosis, and monitoring of a wide range of diseases. Modalities such as magnetic resonance imaging (MRI), computed tomography (CT), ultrasound, and retinal fundus photography provide rich visual data that can be leveraged for clinical decision support. With the increasing volume of medical images being generated, there is a growing need for automated, accurate, and interpretable image analysis tools~\cite{litjens2017survey, erickson2017machine}.

While significant progress has been made using deep learning methods for medical image classification and segmentation, many approaches remain limited in terms of interpretability and generalizability across imaging modalities~\cite{lundervold2019overview}. To address these challenges, recent research has explored the integration of handcrafted features with data-driven models. In particular, gradient-based descriptors and topological data analysis (TDA) have shown promise for capturing both local and global structural information within medical images~\cite{pun2022topological, reininghaus2015stable}.

In this work, we present a novel and generalizable framework that systematically compares \emph{gradient-based} and \emph{topological feature extraction} methods for medical image classification. While our experimental validation focuses on retinal fundus images due to their accessibility and clinical relevance, the proposed methodology is broadly applicable across imaging modalities and disease contexts. We conduct the first comparative analysis of handcrafted feature extraction techniques for retinal disease classification, specifically examining \emph{Histogram of Oriented Gradients} (HOG) and \emph{persistent homology} (PH). Inspired by the foundational work of Dalal and Triggs~\cite{dalal2005histograms}, we adapt HOG descriptors—originally developed for human detection—to the medical imaging domain, extracting $26{,}244$-dimensional gradient-based features from each fundus image. In parallel, we apply PH, a core method in topological data analysis (TDA), to compute 800 topological features using \emph{cubical persistence}, which captures the evolution of structural patterns within medical images. Our findings demonstrate that both feature sets not only perform competitively but also offer complementary insights, laying the groundwork for hybrid image analysis strategies that combine geometric and topological representations to enhance diagnostic accuracy while maintaining interpretability.

We evaluate their performance independently using seven standard machine learning classifiers: logistic regression, random forest, XGBoost, support vector machine, decision tree, $k$-nearest neighbors, and Extra Trees. The experiments are conducted on the publicly available APTOS dataset under two settings: a binary classification task (normal vs.~diabetic retinopathy) and a five-class task reflecting DR severity levels. This first comparative study underscores the strengths and limitations of each feature type, offering insights into how handcrafted features may complement or enhance deep learning architectures in future work.

\medskip

\noindent \textbf{Our contributions.}
\begin{itemize}
    \item We identify and address a gap in the current literature: while both HOG and TDA have been applied independently in medical imaging tasks, no prior study has directly compared their performance on the same dataset or disease classification problem. Our work provides the first such comparison.

    \item We review and contextualize two prominent handcrafted feature extraction methods in medical image analysis: \emph{Histogram of Oriented Gradients} (HOG), which captures local texture and edge information, and \emph{persistent homology} (PH) from topological data analysis (TDA), which encodes global shape and structural connectivity in images.

    \item We apply HOG and TDA to color fundus photographs from the large-scale, publicly available APTOS dataset, targeting two classification tasks: (i) binary classification (normal vs.\ diabetic retinopathy), and (ii) multi-class classification (five DR severity grades).

    \item For each image, we extract $26{,}244$ HOG features and 800 TDA features. These feature sets are independently evaluated using seven classical machine learning models: logistic regression, random forest, XGBoost, support vector machine, decision tree, $k$-nearest neighbors, and Extra Trees, with 10-fold cross-validation.

\item \emph{XGBoost} achieved the best performance across both settings: for binary classification, it reached an accuracy of \textbf{94.29$\pm$1.18\%} (HOG) and \textbf{94.18$\pm$0.92\%} (TDA); for the five-class task, it attained \textbf{74.41$\pm$1.02\%} (HOG) and \textbf{74.69$\pm$1.52\%} (TDA).

    \item Our results highlight that both HOG and TDA are \emph{computationally efficient}, require no data augmentation or pretraining, and produce \emph{interpretable features} that reflect distinct geometric and structural aspects of the images.

    \item We demonstrate that these lightweight, interpretable features are suitable for integration into modern deep learning pipelines, encouraging future hybrid approaches that combine handcrafted features with learned representations to improve diagnostic performance in medical image analysis.
\end{itemize}

\section{Related Work}
\label{sec:related} 

\subsection{Topological Data Analysis (TDA) in Medical Image Analysis} Topological Data Analysis (TDA) is an emerging approach that is increasingly being utilized in medical image analysis, including applications in retinal imaging~\cite{ahmed2023tofi, ahmed2023topological}. Over the past decade, TDA has shown great promise across diverse fields such as image analysis, neurology, cardiology, hepatology, gene-level and single-cell transcriptomics, drug discovery, evolutionary biology, and protein structure analysis. Its strength lies in revealing latent patterns within data, offering new avenues for tasks like image segmentation, object recognition, registration, and reconstruction. In medical imaging, persistent homology (PH)—a key method in TDA—has proven effective in analyzing histopathology slides~\cite{qaiser2019fast,lawson2019persistent,10385822}, fibrin network structures~\cite{berry2020functional}, tumor classification~\cite{crawford2020predicting}, chest X-ray screening~\cite{ahmed2023topo}, retinal disease detection~\cite{ahmed2023tofi, ahmed2023topological, ahmed2025topo}, neuronal morphology studies~\cite{kanari2018topological}, brain artery mapping~\cite{bendich2016persistent}, fMRI analysis~\cite{rieck2020uncovering,stolz2021topological}, and genomic inference~\cite{camara2016inference}.

\smallskip

\subsection{Histogram of Oriented Gradients (HOG) in Medical Image Analysis} 
The Histogram of Oriented Gradients (HOG) is a classical feature descriptor originally introduced for human detection in natural images~\cite{dalal2005histograms}. Due to its ability to capture local edge and gradient information effectively, HOG has been widely adapted for medical image analysis. In recent years, it has been applied to a variety of tasks including retinal disease classification~\cite{ashraf2021hog,al2022retinal}, cervical cancer diagnosis from Pap smear images~\cite{zhang2017hybrid}, breast cancer detection in mammograms~\cite{irshad2013methods}, lung nodule classification in CT scans~\cite{setio2016pulmonary}, and brain tumor segmentation in MRI~\cite{adegun2020hybrid}. HOG features have also proven effective in capturing cell morphology and structure in histopathology and microscopic images~\cite{kumar2017dataset}. Its computational efficiency, interpretability, and robustness to illumination and geometric transformations make it a valuable tool in settings where classical machine learning pipelines are preferred or when combined with deep learning in hybrid systems.

\section{Methodology} \label{sec:Method}

In this study, we explore and compare two distinct feature extraction techniques for retinal fundus images: \emph{persistent homology} (PH), a method rooted in topological data analysis (TDA), and the \emph{Histogram of Oriented Gradients} (HOG), a classical gradient-based descriptor. PH captures the underlying topological structure of image data by identifying homological features—such as connected components, loops, and voids—and tracking their persistence across a filtration scale. For image-specific applications, we employ \emph{cubical persistence}, which is well-suited for grid-structured data like pixel arrays~\cite{wasserman2018topological,chazal2021introduction}. This approach yields topological summaries that are robust to noise and invariant to certain deformations. For background on PH in other contexts, including point clouds and networks, we refer readers to~\cite{dey2022computational,carlsson2021topological}.

In parallel, we apply the HOG method to capture local edge orientation and intensity gradients within the retinal images. Originally introduced for human detection in natural images~\cite{dalal2005histograms}, HOG has proven effective for various computer vision tasks. In our implementation, each retinal image is transformed into a high-dimensional descriptor of size $26{,}244$, encoding rich local texture information.

Rather than combining the two representations, we independently evaluate the performance of models trained using each feature set. This comparative analysis allows us to assess the strengths and limitations of topological versus gradient-based features in the context of retinal disease classification.

\subsection{PH for Image Data: Cubical Persistence}

PH can be summarized as a three-step process. The first step is the \textit{filtration}, where a sequence of simplicial complexes is induced from the data. This step is crucial as it allows integration of domain-specific information. The second step is the creation of \textit{persistence diagrams}, which record the evolution (birth and death) of topological features across the sequence. The final step is \textit{vectorization}, which converts these features into functional or vector forms suitable for machine learning (ML) models. \textit{Note: The following steps are adapted from our previously published work~\cite{ahmed2023tofi}.}

\begin{enumerate}[label=\roman*.]
    \item \textbf{Constructing Filtrations:}The key step in PH is the construction of a sequence of complexes. In image analysis, this is often achieved by generating a nested sequence of binary images (cubical complexes). For a given image $\X$ of resolution $r \times s$, grayscale values $\gamma_{ij}$ of each pixel $\Delta_{ij} \subset \X$ are used.

Given an ordered sequence of grayscale thresholds $t_1 < t_2 < \dots < t_N$, we obtain a sequence $\X_1 \subset \X_2 \subset \dots \subset \X_N$, where $\X_n = \{ \Delta_{ij} \subset \X \mid \gamma_{ij} \leq t_n \}$. This forms a \textit{sublevel filtration}. Alternatively, using $\gamma_{ij} \geq t_n$ yields a \textit{superlevel filtration}.

\begin{figure}[b]
    \centering
    \includegraphics[width=\linewidth]{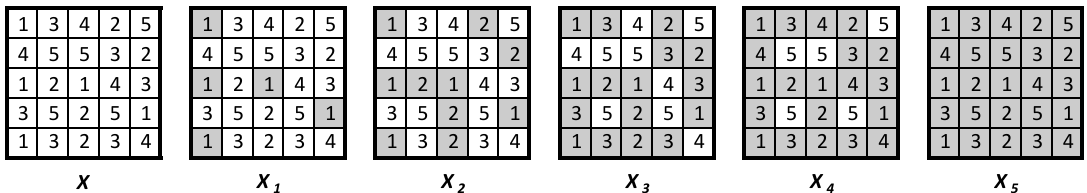}
    \caption{\small {\bf TDA sublevel filtration.} The leftmost figure represents a $5\times 5$ image with pixel values. The sublevel filtration sequence is shown from left to right. (Note: This Fig is reused from our previous published paper ~\cite{ahmed2023tofi})}
    \label{fig:filtration}
\end{figure}

The threshold choice affects the granularity of topological detection. Using $N = 255$ provides fine detail but may yield trivial features. Fewer thresholds risk missing features. Common choices include evenly spaced values or quantiles of the grayscale distribution. In our experiments, $N = 255$ yielded strong results. Sublevel and superlevel filtrations can yield distinct insights. However, for cubical complexes in image data, Alexander Duality~\cite{hatcher2002algebraic} ensures they produce nearly equivalent information across complementary dimensions. Thus, using both dimensions $k=0$ and $k=1$ captures the complete information.
    \item  \textbf{Persistence Diagrams:} Given the filtration $\X_1 \subset \X_2 \subset \dots \subset \X_N$, persistence diagrams (PDs) summarize the topological changes by recording birth and death of features. These diagrams contain tuples $(b_\sigma, d_\sigma)$ representing when a feature $\sigma$ appears and disappears across the filtration.

Formally, 
\[
\mathrm{PD}_k(\X) = \{(b_\sigma, d_\sigma) \mid \sigma \in H_k(\wh{\X}_i), \ b_\sigma \leq i < d_\sigma\},
\]
where $H_k(\wh{\X}_i)$ denotes the $k$-th homology group of complex $\wh{\X}_i$~\cite{hatcher2002algebraic}.

For 2D image analysis, only $k=0$ (connected components) and $k=1$ (holes or loops) are relevant. For example, if a loop appears at $\X_3$ and disappears at $\X_7$, the tuple $(3,7)$ appears in $\PD_1(\X)$. In Figure~\ref{fig:filtration}, we have:
\begin{align*}
\PD_0(\X) &= \{(1,\infty), (1,2), (1,3), (1,3), (1,4), (2,3)\}, \\
\PD_1(\X) &= \{(3,5), (3,5), (4,5)\}.
\end{align*}

Software libraries for this include Giotto-TDA~\cite{tauzin2021giotto} for image data and other tools listed in~\cite{otter2017roadmap}.
    \item  \textbf{Vectorization:}
PDs as tuple collections are not directly usable for ML. We apply \textit{vectorization}, such as using \textit{Betti functions}, which count "alive" features at each threshold. Define:
\[
\vec{\beta}_k(\X) = [\beta_k(t_1)\ \beta_k(t_2)\ \dots\ \beta_k(t_N)],
\]
where $\beta_0(t_n)$ is the number of components and $\beta_1(t_n)$ is the number of holes at threshold $t_n$.

From Figure~\ref{fig:filtration}, we compute:
\[
\vec{\beta}_0(\X) = [5\ 4\ 2\ 1\ 1], \quad \vec{\beta}_1(\X) = [0\ 0\ 2\ 3\ 0].
\]

In our experiment, we aggregate Betti vectors into $N=100$ bins to reduce dimensionality. Thus, for each channel, we extract 100-dimensional $\beta_0$ and $\beta_1$ vectors. These are used as features for ML classification. Although other methods exist (Persistence Images, Silhouettes~\cite{dey2022computational}), we favor Betti vectors for interpretability.

\end{enumerate}




\begin{figure}[t]
    \centering
    \includegraphics[width=\linewidth]{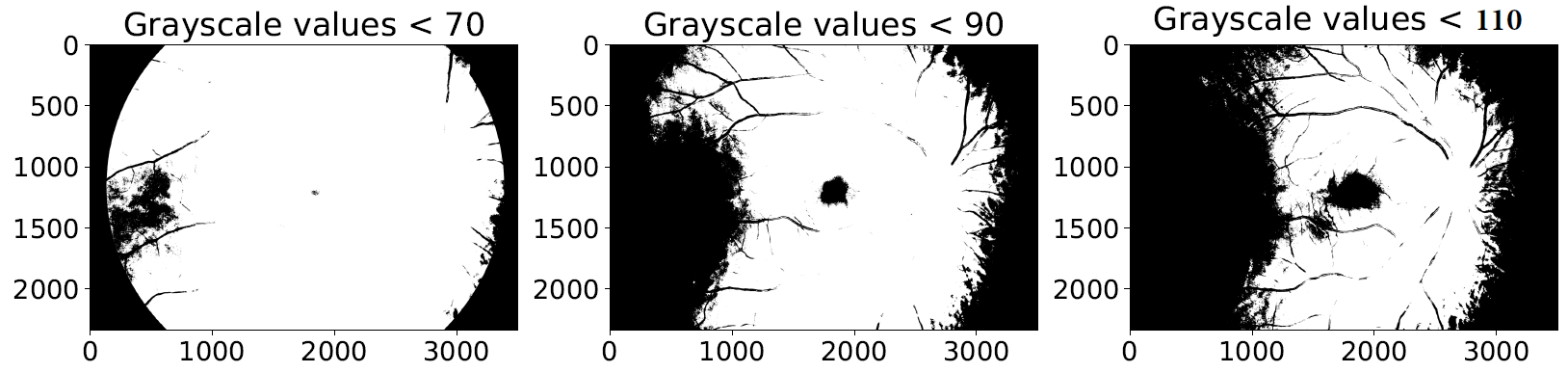}
    \caption{\small {\bf Sublevel filtration.} Binary images ${\X_{70}},{\X_{90}},{\X_{110}}$ obtained from a fundus image for threshold values $70,90,110$. (Note: This Fig is reused from our previous published paper ~\cite{ahmed2023tofi})}
    \label{fig:filtration2}
\end{figure}

\subsubsection{Color Channels for Retinal Images}

Retinal image quality is vital for reliable diagnoses. Different color spaces affect the feature extraction process. We use RGB and grayscale channels to construct filtrations, each producing different topological patterns.

\begin{figure*}[t!]
    \centering
    \includegraphics[width=\linewidth]{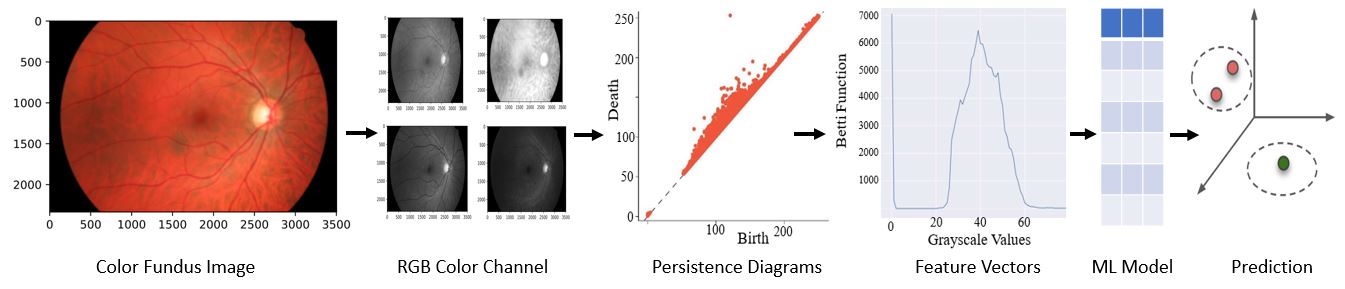}
    \caption{\footnotesize {\bf Flowchart of TDA-ML model pipeline:} Starting from color fundus images, we extract grayscale and RGB channels. Persistence diagrams are computed from these color spaces, from which 100-dimensional Betti-0 and Betti-1 vectors are derived for each channel. These topological feature vectors are concatenated and used as input to machine learning classifiers such as Random Forest (RF), XGBoost, and k-Nearest Neighbors (kNN) to achieve accurate retinal image classification. (Note: This Fig is reused from our previous published paper ~\cite{ahmed2023tofi})}
    \label{fig:tda-flowchart}
\end{figure*}

\subsubsection{Extracting Topological Features} \label{sec:Topological_features}

We now describe the full pipeline of our model and its interpretability.

Given a fundus image $\X$, we derive grayscale and RGB channels: $\mathbf{g}(i,j)$ (grayscale), $\mathbf{R}(i,j)$, $\mathbf{G}(i,j)$, and $\mathbf{B}(i,j)$. Each defines a color function over pixels $\Delta_{ij} \subset \X$.

Using $N=255$ grayscale thresholds, we construct sublevel filtrations and compute persistence diagrams $PD_k(\X)$ for $k=0,1$. As illustrated in Figure~\ref{fig:filtration2}, each filtration yields a sequence of binary images, where black pixels have values below the current threshold.

Persistent diagrams are vectorized using Betti functions. From each image, we obtain 100-dimensional vectors for $\beta_0$ and $\beta_1$ across four channels (Grayscale, R, G, B), yielding a total of 800-dimensional feature vectors. These topological features are then used as input to machine learning models for retinal image classification.

\begin{figure}[t!] 
	\centering
	\subfloat[\scriptsize Color retinal fundus image \label{fig:Color-HOG img}]{%
		\includegraphics[width=0.32\linewidth]{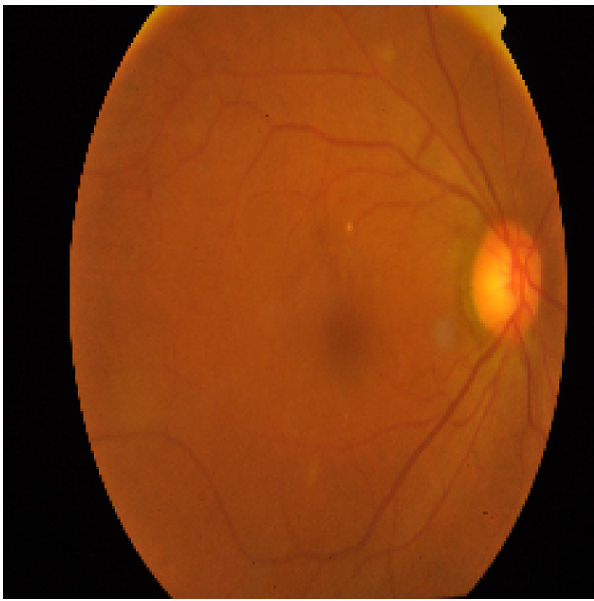}}
	\hfill
	\subfloat[\scriptsize Grayscale retinal fundus image\label{fig:color img}]{%
		\includegraphics[width=0.32\linewidth]{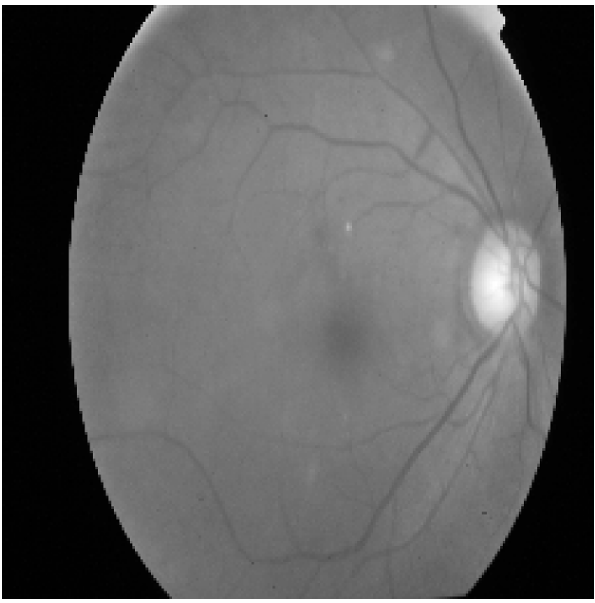}}
	\hfill
	\subfloat[\scriptsize HOG visualization \label{fig:HOG img}]{%
		\includegraphics[width=0.32\linewidth]{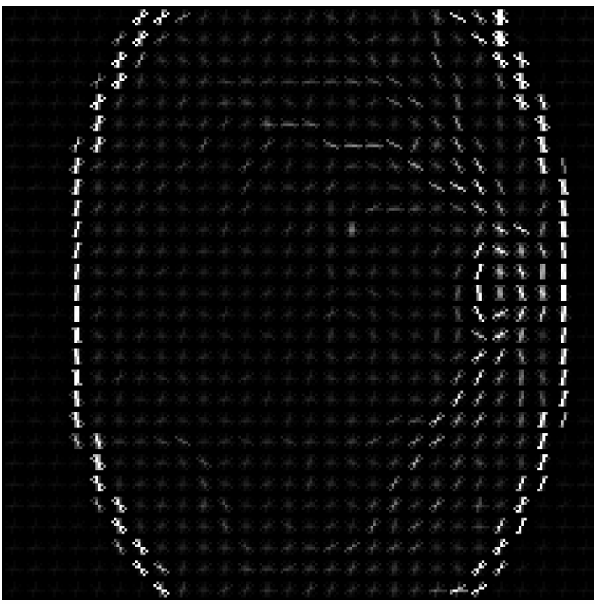}}
	\caption{\footnotesize Illustration of the preprocessing pipeline applied to a fundus image. From left to right: (a) Original color image, (b) grayscale conversion used for HOG computation, and (c) corresponding Histogram of Oriented Gradients (HOG) visualization highlighting edge and texture features.}
	\label{fig:HOG-visualization}
\end{figure}
\subsection{HOG Feature Extraction from Retinal Images}

In this study, we employed the Histogram of Oriented Gradients (HOG) as a feature extraction technique to capture the texture and structural characteristics of retinal fundus images. HOG is a widely-used descriptor for object detection and image classification due to its ability to encode gradient orientation distributions, which are particularly useful in representing local shape and edge information.

The feature extraction pipeline began with data acquisition, denoted as $\mathcal{I} = \{I_1, I_2, \ldots, I_n\}$, where each $I_k$ represents a color fundus image. Each image was resized to a uniform spatial resolution of $224 \times 224$ pixels to ensure consistency across the dataset. This resizing can be denoted as:
\[
I_k' = \text{Resize}(I_k, 224 \times 224)
\]

Following resizing, each RGB image $I_k'$ was converted to a grayscale image $G_k$ using a luminance-preserving transformation:
\[
G_k(x, y) = 0.299 \cdot R(x, y) + 0.587 \cdot G(x, y) + 0.114 \cdot B(x, y)
\]
This grayscale conversion is essential since HOG operates on intensity gradients, and color information does not significantly enhance gradient-based descriptors.

Once converted, the HOG descriptor was computed on each grayscale image. This involved calculating the gradient magnitude and orientation at each pixel using discrete derivative masks. The image was then partitioned into small spatial regions known as \emph{cells}, each of size $8 \times 8$ pixels. Within each cell, a histogram of gradient directions was computed using 9 orientation bins (covering $0^\circ$ to $180^\circ$). These histograms describe the distribution of edge directions within the local area.

To account for changes in illumination and contrast, these local histograms were grouped into overlapping \emph{blocks} of $2 \times 2$ cells. Each block's histograms were concatenated and normalized using the L2-Hys norm to produce a robust local feature vector. Mathematically, this can be expressed as:
\[
F_k = \text{HOG}(G_k; o=9, p=8 \times 8, b=2 \times 2)
\]
where $F_k \in \mathbb{R}^d$ is the resulting high-dimensional HOG descriptor for image $k$, and $d$ denotes the length of the flattened feature vector. The value of $d$ depends on the total number of cells and blocks given the fixed image resolution. In our experiment, $d=26{,}244$, so $F_k \in \mathbb{R}^{26{,}244}$; that is, from each grayscale image, we extract a $26{,}244-$dimensional vector. 

This process was repeated for all $n$ images in the dataset, resulting in a feature matrix:
\[
\mathbf{F} = 
\begin{bmatrix}
F_1 \\
F_2 \\
\vdots \\
F_n
\end{bmatrix}
\in \mathbb{R}^{n \times d}
\]

The matrix $\mathbf{F}$ was then stored in tabular format using a DataFrame structure and exported as a CSV file named \texttt{hog\_features.csv}. This stored feature set served as a structured and compressed representation of the input images, which was later used for machine learning models in the classification tasks.

Overall, this HOG-based feature extraction step aimed to leverage handcrafted descriptors that capture geometrical and textural cues in retinal images, providing a complementary signal to the retinal fundus image classification.

\begin{figure*}[t!]
    \centering
    \includegraphics[width=\linewidth]{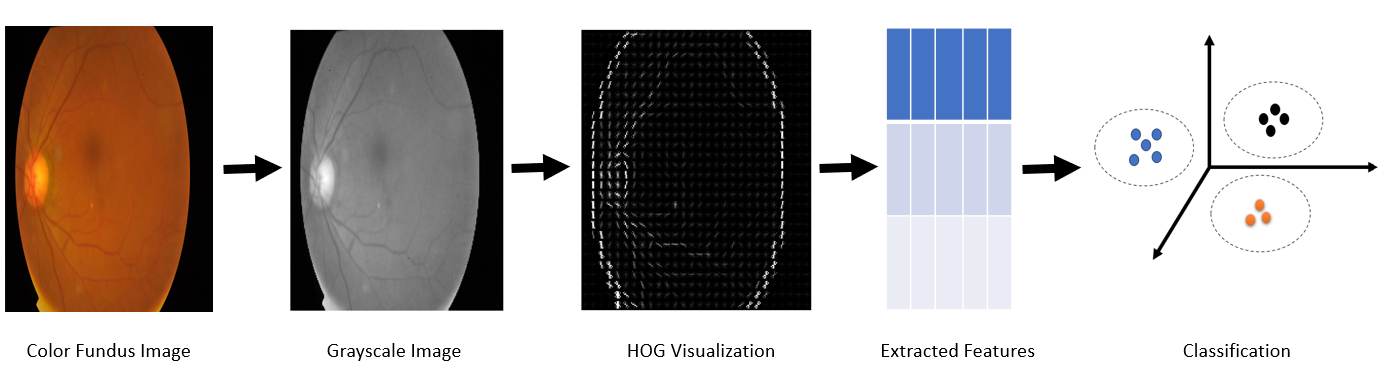}
    \caption{\footnotesize {\bf HOG-ML Model Pipeline:} The flowchart illustrates the complete processing pipeline for retinal image classification using HOG features. Starting with color fundus images, grayscale conversion is performed, followed by Histogram of Oriented Gradients (HOG) computation and visualization. From each grayscale image, a total of 26,244 HOG features are extracted. These feature vectors are then used as input to machine learning classifiers such as Random Forest (RF), XGBoost, and k-Nearest Neighbors (kNN) to achieve accurate classification of retinal images.}
    \label{fig:hog-flowchart}
\end{figure*}

\section{Machine Learning Models}

To classify the extracted features from Topological Data Analysis (TDA) and Histogram of Oriented Gradients (HOG), we employed seven classical machine learning models widely used for binary classification tasks: Logistic Regression, Random Forest, XGBoost, K-Nearest Neighbors (KNN), Decision Tree, Support Vector Machine (SVM), and Extra Trees. Below, we briefly describe the theoretical foundation and mathematical formulation of each model.

\subsection{Logistic Regression}

Logistic Regression assumes that the \emph{log-odds} of the positive class are a linear function of the input features.  For a sample $\mathbf{x}\in\mathbb{R}^d$ the model estimates
\[
P(y=1\mid\mathbf{x}) \;=\; \sigma(\mathbf{w}^{\!\top}\mathbf{x} + b)
   \;=\; \frac{1}{1 + e^{-(\mathbf{w}^{\!\top}\mathbf{x} + b)}} ,
\]
where $\sigma$ denotes the sigmoid.  Parameters $(\mathbf{w},b)$ are obtained by minimizing the regularised cross-entropy
\[
\mathcal{L}(\mathbf{w},b)
  \;=\;
  -\frac1N\sum_{i=1}^N \bigl[y_i\log p_i + (1-y_i)\log(1-p_i)\bigr]
  \;+\; \lambda\lVert\mathbf{w}\rVert_2^2 ,
\]
so the decision boundary $\mathbf{w}^{\!\top}\mathbf{x}+b=0$ is linear and the predicted class is $\hat{y}=1$ when $p_i\!\ge\!0.5$.

\subsection{Random Forest}

A Random Forest is an ensemble of $T$ independent decision trees, each trained on a bootstrap replicate of the data and, at every split, constrained to search only a random subset of $m<d$ features.  For classification the forest outputs the majority vote
\[
\hat{y}
  \;=\;\operatorname{mode}\{h_t(\mathbf{x})\}_{t=1}^T ,
\]
with $h_t$ the prediction of the $t$-th tree.  Bagging decorrelates trees, and the random feature sub-sampling injects additional variance, jointly reducing the overall variance without increasing bias, thus limiting overfitting.

\subsection{Extremely Randomized Trees (Extra Trees)}

Extra Trees keep the global forest architecture but push randomness further: trees are grown on the entire training set (no bootstrapping) and, when a node needs to split, each candidate feature is paired with a \emph{completely random} cut-point drawn uniformly in its range; the best of these randomly proposed splits (according to the impurity criterion, e.g.\ Gini) is chosen.  The high variance of individual trees is offset by averaging, yielding a very low-variance ensemble that is often faster to train than a Random Forest.

\subsection{Decision Tree}

A Decision Tree partitions the feature space recursively.  At each node it selects the feature–threshold pair that maximises the purity gain.  Using Gini impurity the gain at node $s$ is
\[
\Delta G_s
  \;=\;
  G(\text{parent})\;-\;\sum_{c\in\{L,R\}} \frac{n_c}{n_s}\,G(c),\quad
  G(\cdot)=1-\!\!\sum_{k\in\{0,1\}}\!\!p_k^{\,2}.
\]
The tree keeps splitting until stopping criteria (depth, minimum leaf size, or purity) are met, then assigns the majority class of each leaf to unseen samples.

\subsection{K-Nearest Neighbors}

KNN is a lazy, non-parametric method.  Given a query $\mathbf{x}$, let $\mathcal N_k(\mathbf{x})$ be the indices of the $k$ closest training points under a distance $d(\cdot,\cdot)$ (Euclidean in our experiments).  The predicted label is
\[
\hat{y}
  \;=\;
  \operatorname{mode}\bigl\{y_i : i\in\mathcal N_k(\mathbf{x})\bigr\},
\]
and an optional posterior probability estimate is $P(y=1)=\frac{1}{k}\sum_{i\in\mathcal N_k(\mathbf{x})}y_i$.  Complexity is shifted to inference ($\mathcal O(kN)$ naïvely), while the model can approximate an arbitrarily complex decision boundary given sufficient data.

\subsection{Support Vector Machine}

For linearly separable data the hard-margin SVM finds the hyperplane
$\mathbf{w}^{\!\top}\mathbf{x}+b=0$ maximising the margin $2/\lVert\mathbf{w}\rVert_2$.  In practice we solve the soft-margin primal
\[
\min_{\mathbf{w},b,\xi}\;
        \frac{1}{2}\lVert\mathbf{w}\rVert_2^2
        + C\!\sum_{i=1}^N \xi_i
\quad
\text{s.t. }\;
y_i(\mathbf{w}^{\!\top}\phi(\mathbf{x}_i) + b)\ge 1-\xi_i,\;
\xi_i\ge0,
\]
where $\phi$ is an implicit mapping defined by a kernel $K(\mathbf{x},\mathbf{x}')=\langle\phi(\mathbf{x}),\phi(\mathbf{x}')\rangle$.  We use the RBF kernel to capture non-linear structure, and classification is $\hat{y}=\operatorname{sign}(\sum_i \alpha_i y_i K(\mathbf{x},\mathbf{x}_i)+b)$ with $\alpha_i$ the learned dual coefficients.

\subsection{XGBoost}

XGBoost builds an additive model of $T$ regression trees,
$\hat{y}_i^{(T)}=\sum_{t=1}^{T}f_t(\mathbf{x}_i)$, and optimises
\[
\mathcal{L}^{(T)}
  \;=\;
  \sum_{i=1}^N l\bigl(y_i, \hat{y}_i^{(T)}\bigr)
  + \sum_{t=1}^{T}\Omega(f_t),\quad
  \Omega(f)=\gamma T_f + \tfrac12\lambda\lVert\mathbf{w}_f\rVert_2^{2},
\]
where $l$ is the logistic loss for binary classification, $T_f$ counts leaves of tree $f$, and $\mathbf{w}_f$ stores its leaf scores.  At round $t$ the algorithm fits $f_t$ to the negative gradient of the loss (the residual) and prunes splits whose gain falls below $\gamma$.  Shrinkage, column sub-sampling, and second-order gradient information together make XGBoost highly accurate, fast, and robust to overfitting.

\section{Experiments}
\label{sec:experiments}

\subsection{Datasets} \label{sec:datasets}

\noindent {\bf APTOS 2019 dataset} was used for a Kaggle competition on DR diagnosis~\cite{aptos2019}. The images have varying resolutions, ranging from $474\times 358$ to $3388\times 2588$. APTOS stands for Asia Pacific Tele-Ophthalmology Society, and the dataset was provided by Aravind Eye Hospital in India. 
In this dataset, fundus images are graded manually on a scale of 0 to 4 (0: no DR; 1: mild; 2: moderate; 3: severe; and 4: proliferative DR) to indicate different severity levels. The number of images in these classes are respectively 1805, 370, 999, 193, and 295. In the binary setting, class 0 is defined as the normal group, and the remaining classes (1-4) are defines as DR group which gives a split 1805:1857. The total number of training and test samples in the dataset were 3662 and 1928 respectively. However, the labels for the test samples were not released after the competition, so like other references, we used the available 3662 fundus images with labels. We report our results on a binary and 5-class classification setting. In a binary setting,  fundus  images with grades 1,2,3, and 4 are identified as DR group, and grade 0 images as the normal group.

\subsection{Experimental Setup} \

\noindent {\bf Training–Test Split:} All experiments were conducted using 10-fold cross-validation to ensure robust and reliable performance evaluation across all machine learning models. For each fold, the model was trained on 90\% of the data and tested on the remaining 10\%. We report the mean performance of each model across the 10 folds for all evaluation metrics—accuracy, precision, recall, F1-score, and AUC—along with their corresponding standard deviations to reflect variability and stability.
 
\smallskip

\noindent {\bf No Data Augmentation:} Our Topo-ML model are using topological feature vectors, and our feature extraction method is invariant under rotation, flipping and other common data augmentation techniques. Hence, we do not use any type of data augmentation or pre-processing to increase the size of training data for ML models. This makes our model computationally very efficient, and highly robust against small alterations and the noise in the image.

\smallskip

\noindent {\bf ML Model Hyperparameters:} All seven machine learning models—Logistic Regression, Random Forest, XGBoost, Support Vector Machine (SVM), Decision Tree, $k$-Nearest Neighbors (KNN), and Extra Trees—were trained using their default hyperparameter settings as provided by the \texttt{scikit-learn} and \texttt{xgboost} libraries. No manual tuning or optimization was performed to maintain consistency and fairness in comparison across feature types. Each model was evaluated using 10-fold cross-validation, and we report the mean and standard deviation for all performance metrics across the folds.

\smallskip

\noindent {\bf Computational Complexity \& Implementation:} While for high dimensional data PH calculation is computationally expensive~\cite{otter2017roadmap}, for image data, it is highly efficient.  For 2D images, PH  has time complexity of $\mathcal{O}(|\mathcal{P}|^r)$ where $r\sim 2.37$ and $|\mathcal{P}|$ is the total number of pixels~\cite{milosavljevic2011zigzag}. In other words, PH computation increases almost quadratic with the resolution. The remaining processes (vectorization, RF) are negligible compared to PH step. We used Giotto-TDA~\cite{tauzin2020giottotda}  to obtain persistence diagrams, and Betti functions. HOG feature extraction is relatively fast, as the input images are resized to $256 \times 256$ pixels. We used Jupyter notebook as an IDE for writing the code in Python 3. Our code is available at the following link~\footnote{ \url{https://github.com/FaisalAhmed77/TDA-HOG-Comparision/tree/main}}. 

\noindent {\bf Runtime:} We conducted all our experiments utilizing a personal laptop equipped with an Intel(R) Core(TM) i7-8565U processor running at 1.80GHz and boasting 16 GB of RAM. In both of our models, the most time-intensive phase is the extraction of topological features. In contrast, the subsequent tasks of machine learning classification and executing deep learning models are relatively insignificant. This is mainly because we did not employ any data augmentation.
For the most extensive dataset in our study, APTOS (consisting of 3662 high-resolution images), the entire process, including topological feature extraction, model training, and obtaining accuracy results, consumed a total of 43.7 hours. The runtime for smaller datasets with lower resolutions, used for topological feature extraction, is considerably shorter. HOG feature extraction is relatively fast, as the input images are resized to $256 \times 256$ pixels.
 It's also worth considering that when using a server, as opposed to a personal laptop, the runtime for such datasets would be considerably shorter.

\vspace{-.1in}

\begin{figure*}[t!]
	\centering
	\subfloat[\scriptsize Logistic Regression – AUC using TDA Features\label{fig:LR-tda img}]{%
		\includegraphics[width=0.32\linewidth]{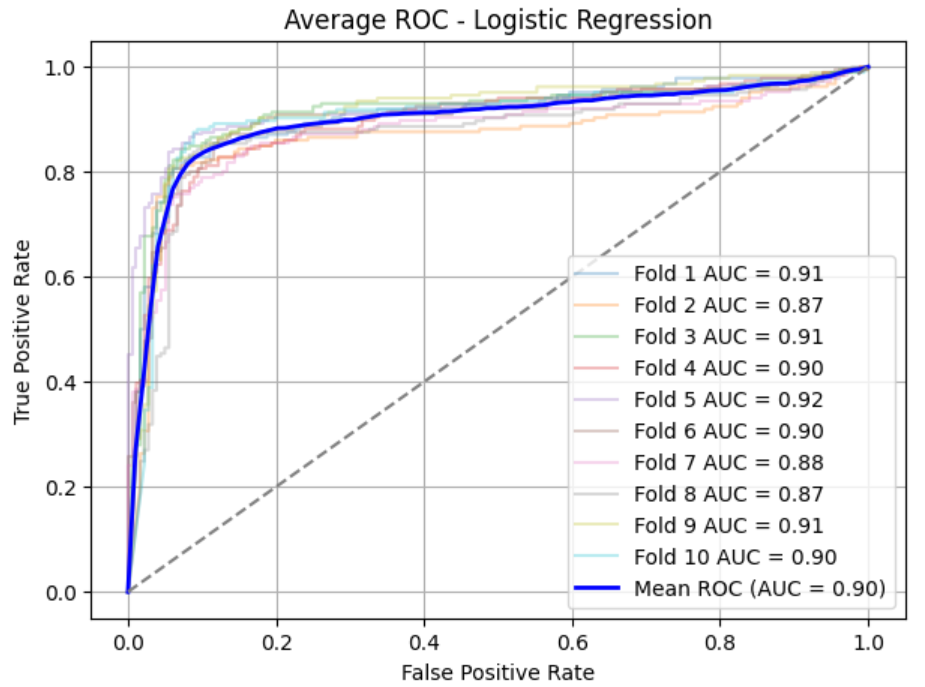}}
	\hfill
	\subfloat[\scriptsize Random Forest – AUC using TDA Features\label{fig:RF-tda img}]{%
		\includegraphics[width=0.32\linewidth]{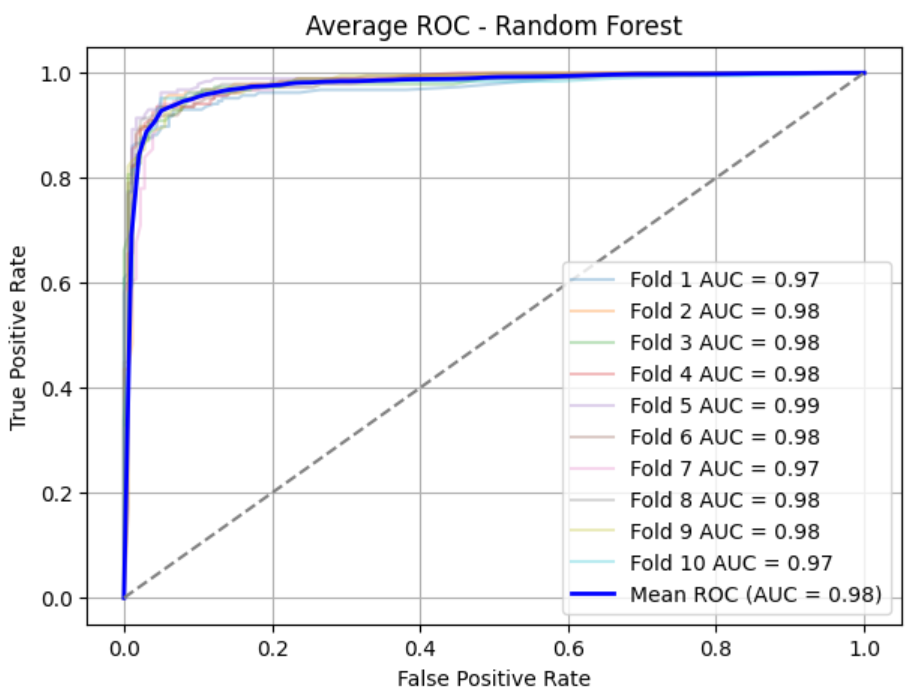}}
	\hfill
	\subfloat[\scriptsize XGBoost – AUC using TDA Features\label{fig:XGB-tda img}]{%
		\includegraphics[width=0.32\linewidth]{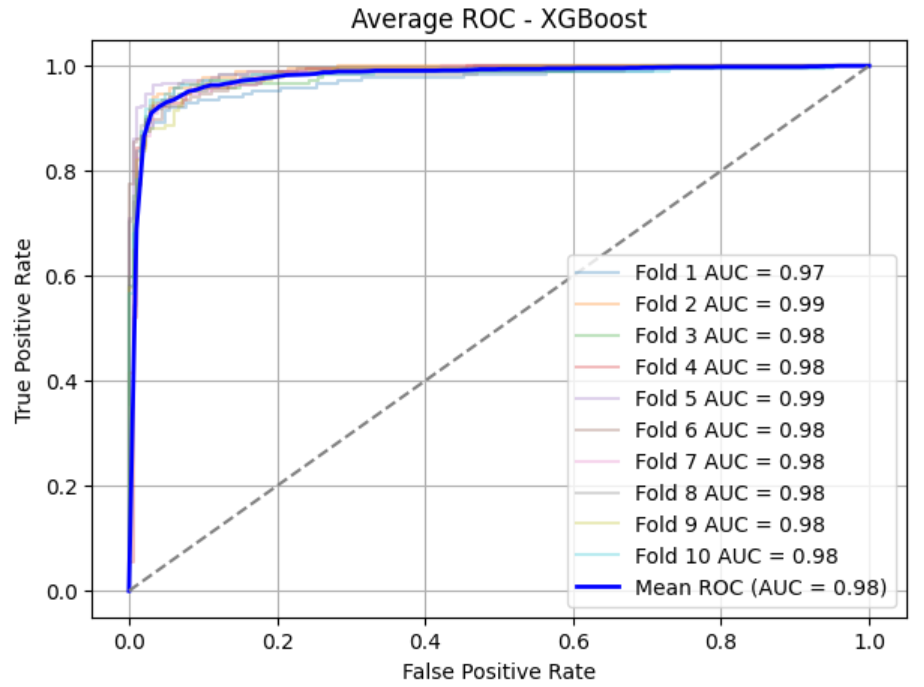}}
	\hfill
	\subfloat[\scriptsize K-Nearest Neighbors – AUC using TDA Features\label{fig:KNN-tda img}]{%
		\includegraphics[width=0.32\linewidth]{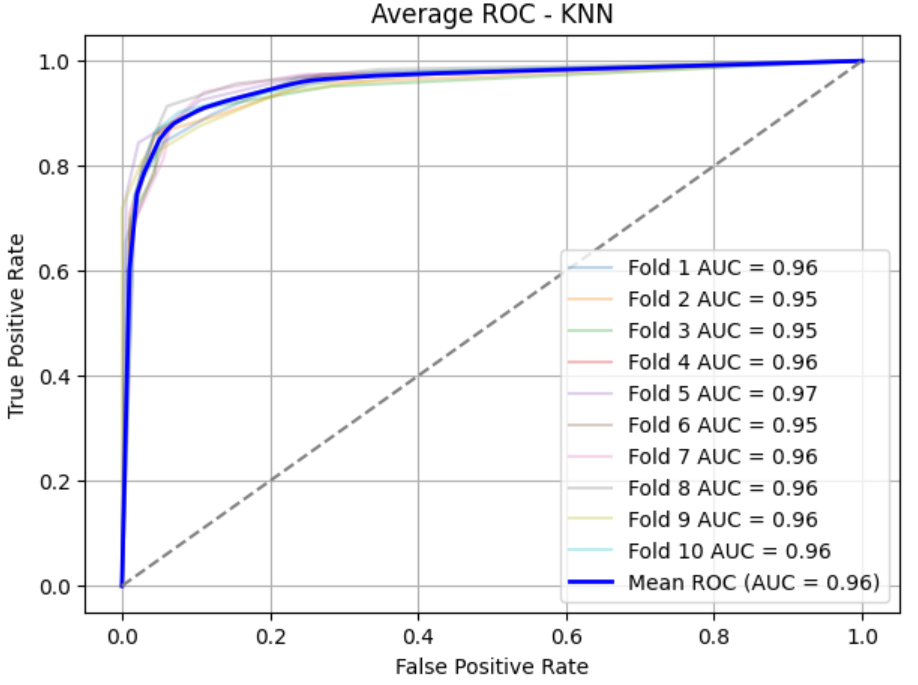}}
	\hfill
	\subfloat[\scriptsize Support Vector Machine – AUC using TDA Features\label{fig:SVM-tda img}]{%
		\includegraphics[width=0.32\linewidth]{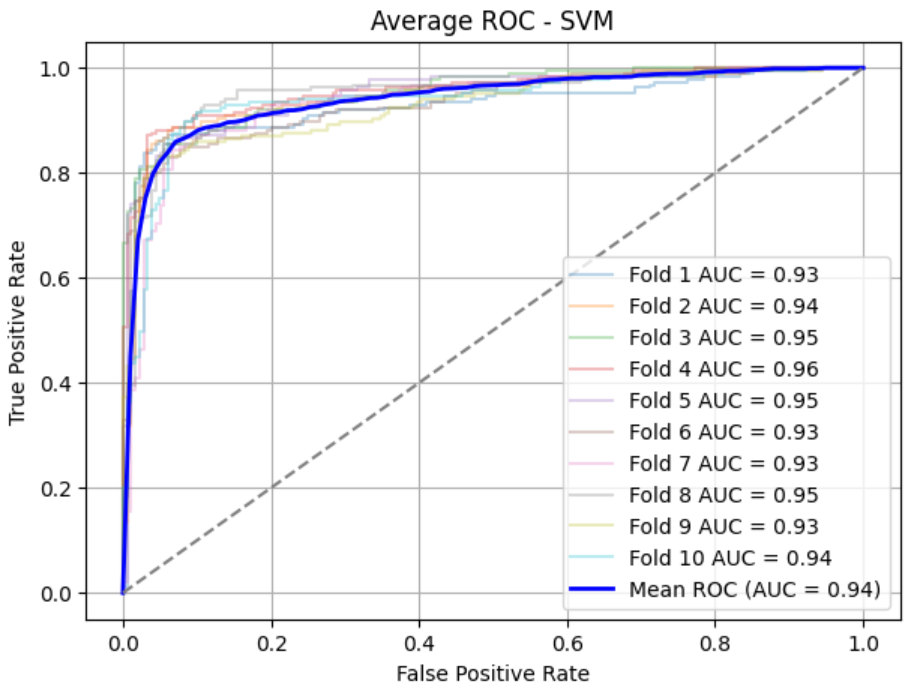}}
	\hfill
	\subfloat[\scriptsize Extra Trees – AUC using TDA Features\label{fig:EXT-tda img}]{%
		\includegraphics[width=0.32\linewidth]{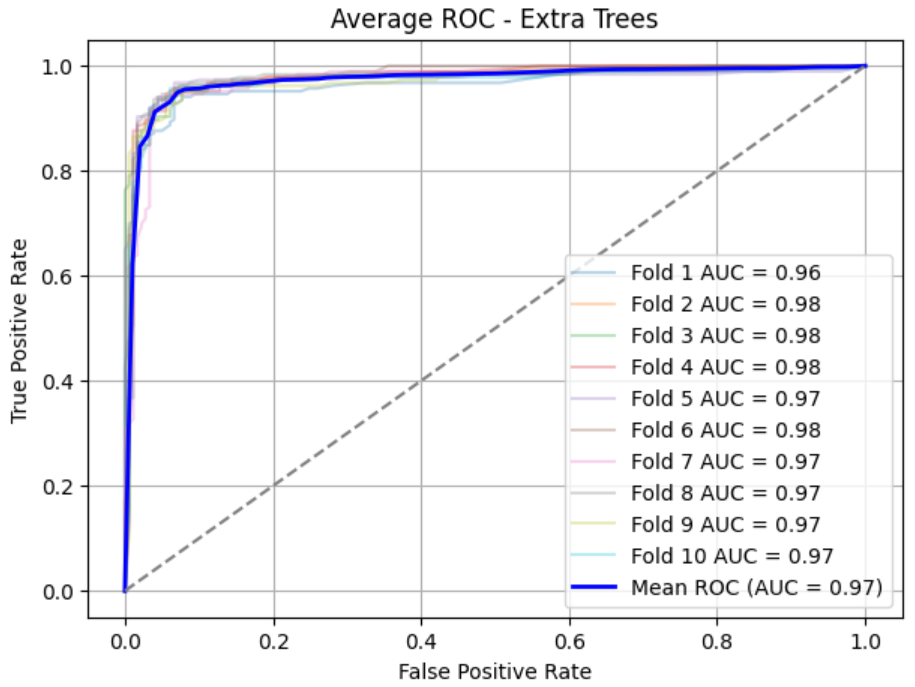}}
	\caption{\footnotesize Comparison of AUC performance for various classifiers using TDA-derived features on the APTOS dataset (binary classification setting).}
	\label{fig:TDA-AUC}
\end{figure*}

\begin{figure*}[t!]
	\centering
	\subfloat[\scriptsize Logistic Regression – Confusion Matrix using TDA Features\label{fig:LR-tda cm}]{%
		\includegraphics[width=0.32\linewidth]{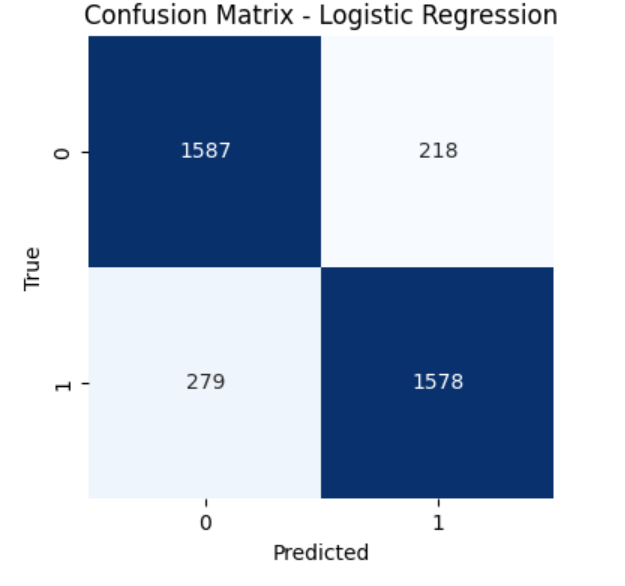}}
	\hfill
	\subfloat[\scriptsize Random Forest – Confusion Matrix using TDA Features\label{fig:RF-tda cm}]{%
		\includegraphics[width=0.32\linewidth]{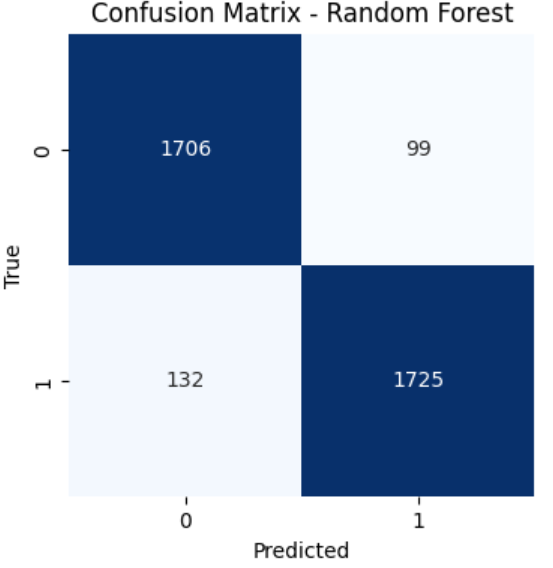}}
	\hfill
	\subfloat[\scriptsize XGBoost – Confusion Matrix using TDA Features\label{fig:XGB-tda cm}]{%
		\includegraphics[width=0.32\linewidth]{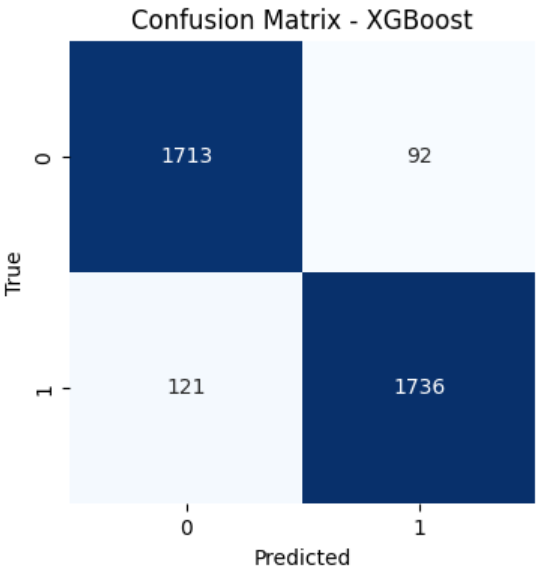}}
	\hfill
	\subfloat[\scriptsize K-Nearest Neighbors – Confusion Matrix using TDA Features\label{fig:KNN-tda cm}]{%
		\includegraphics[width=0.32\linewidth]{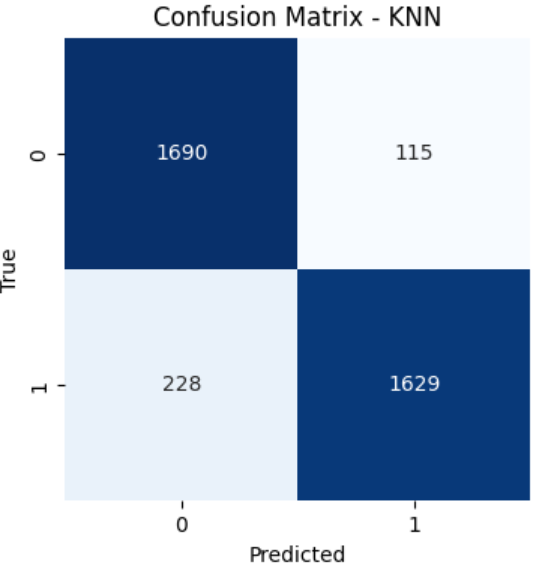}}
	\hfill
	\subfloat[\scriptsize Support Vector Machine – Confusion Matrix using TDA Features\label{fig:SVM-tda cm}]{%
		\includegraphics[width=0.32\linewidth]{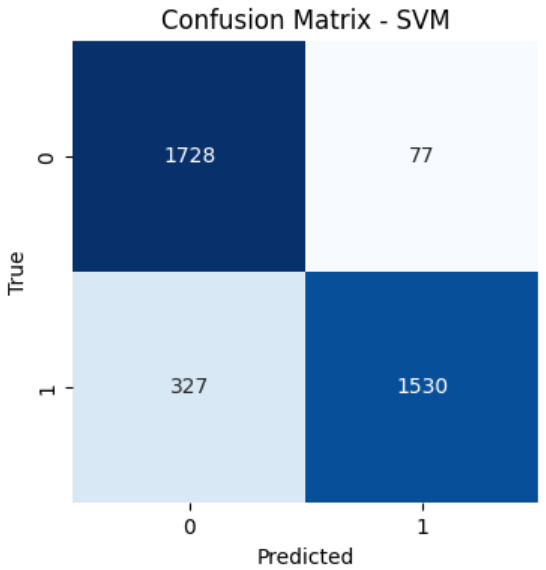}}
	\hfill
	\subfloat[\scriptsize Extra Trees – Confusion Matrix using TDA Features\label{fig:EXT-tda cm}]{%
		\includegraphics[width=0.32\linewidth]{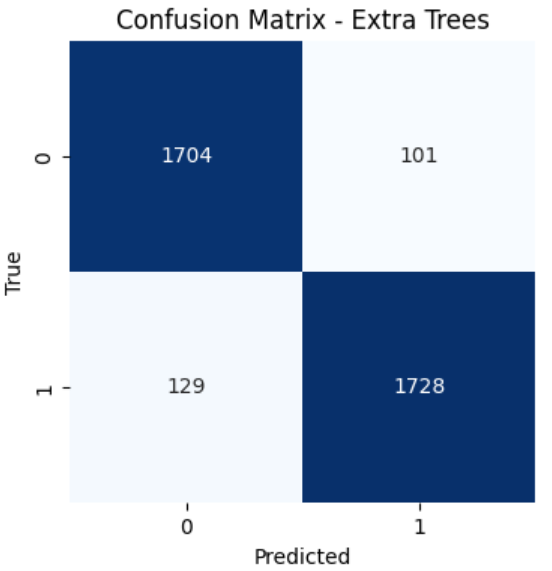}}
	\caption{\footnotesize Confusion matrix comparison of various classifiers trained on TDA-derived features using the APTOS dataset (binary classification setting).}
	\label{fig:TDA-CM}
\end{figure*}

\section{Results} \label{sec:accuracy} 

\subsection{Binary Classification: Normal vs DR}

We first evaluate the performance of both TDA and HOG features in a binary classification setting where the goal is to distinguish normal retinal fundus images from those affected by diabetic retinopathy (DR). The classification experiments are conducted using seven traditional machine learning models, and performance is reported in terms of accuracy, precision, recall, F1-score, and area under the ROC curve (AUC). Tables~\ref{tab:tda_binary} and~\ref{tab:hog_binary} present the average performance metrics (with standard deviations) across 10-fold cross-validation.

\subsubsection*{Comparative Analysis on binary setting: TDA vs HOG Features}

In the binary classification setting (Normal vs DR), both TDA and HOG feature sets yield strong performance across a range of traditional machine learning classifiers. However, HOG features generally outperform TDA features in terms of average accuracy, precision, recall, and F1-score. For instance, Logistic Regression achieves an accuracy of 93.97\% with HOG compared to 86.43\% with TDA. Similarly, Support Vector Machine (SVM) performs better with HOG (94.05\% accuracy) than with TDA (88.97\% accuracy), showing a notable improvement in recall and F1-score.

While both feature types enable ensemble models such as XGBoost and Random Forest to perform exceptionally well, the differences are marginal. XGBoost achieves nearly identical accuracy using TDA (94.18\%) and HOG (94.29\%), suggesting that both feature types are competitive under tree-based classifiers. Notably, TDA features lead to slightly higher AUC scores in Random Forest and Extra Trees (98.0\% and 97.0\%) compared to HOG (both 97.0\%), indicating stronger class separability in some models.

Overall, HOG features tend to provide higher performance consistency across models, especially in linear and kernel-based methods, while TDA offers competitive results and better interpretability. This comparative analysis highlights the complementary strengths of both methods for fundus image classification tasks.

\begin{figure*}[t!]
	\centering
	\subfloat[\scriptsize Logistic Regression – AUC using HOG Features\label{fig:LR-hog img}]{%
		\includegraphics[width=0.32\linewidth]{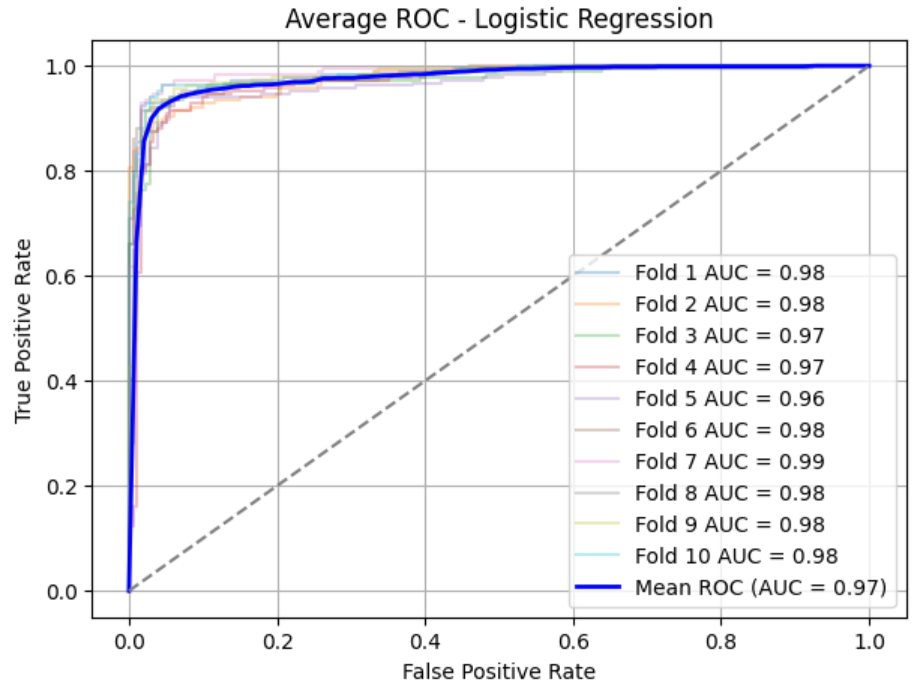}}
	\hfill
	\subfloat[\scriptsize Random Forest – AUC using HOG Features\label{fig:RF-hog img}]{%
		\includegraphics[width=0.32\linewidth]{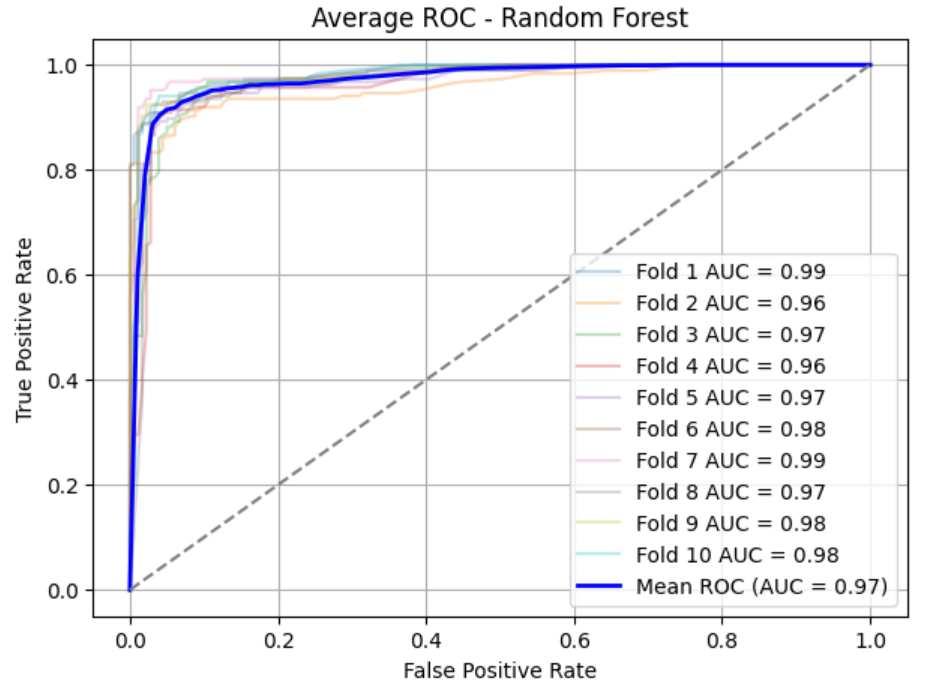}}
	\hfill
	\subfloat[\scriptsize XGBoost – AUC using HOG Features\label{fig:XGB-hog img}]{%
		\includegraphics[width=0.32\linewidth]{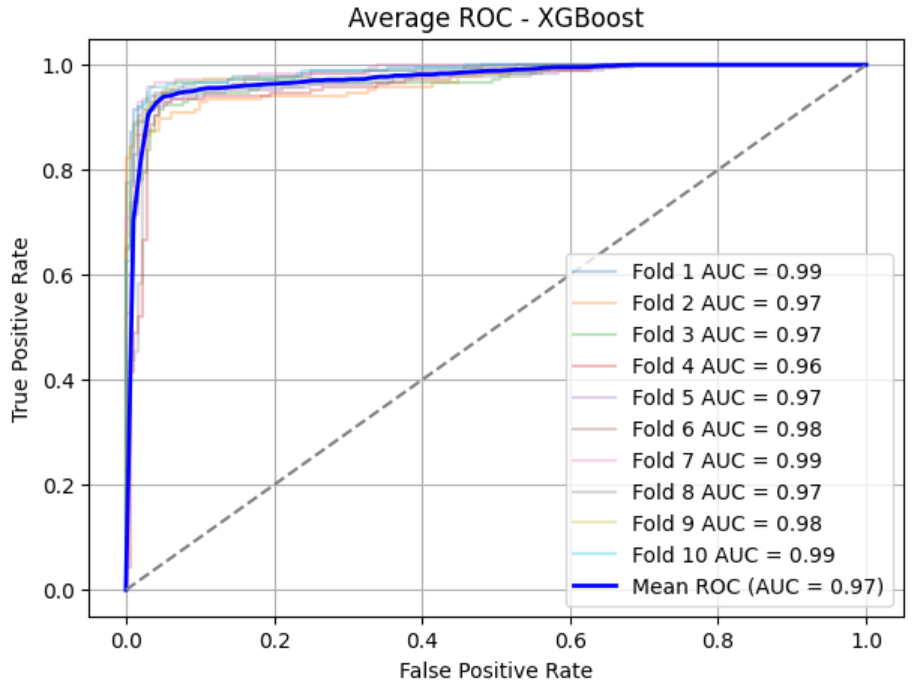}}
	\hfill
	\subfloat[\scriptsize K-Nearest Neighbors – AUC using HOG Features\label{fig:KNN-hog img}]{%
		\includegraphics[width=0.32\linewidth]{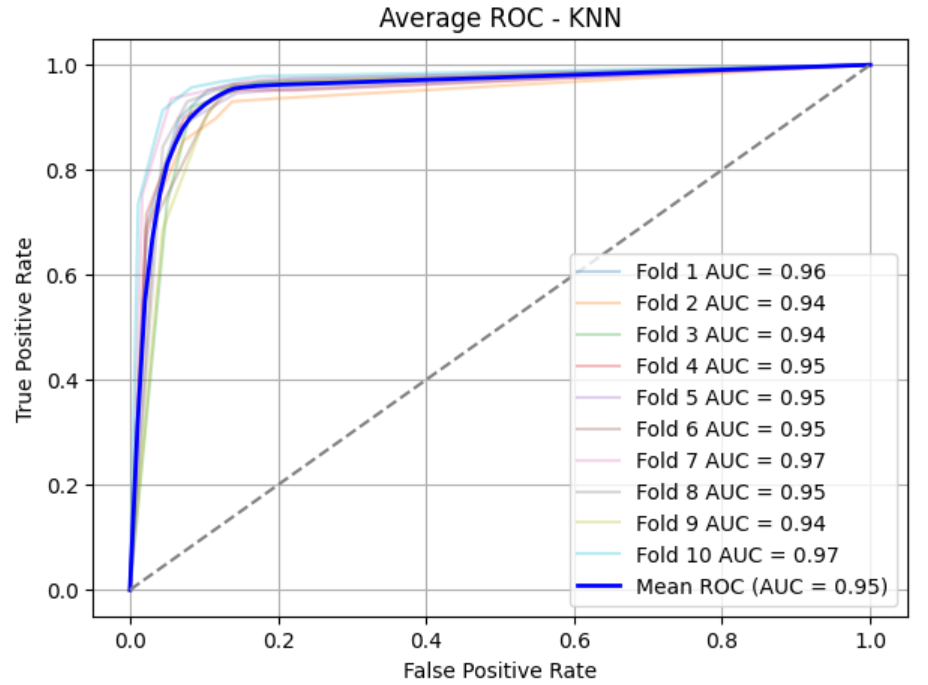}}
	\hfill
	\subfloat[\scriptsize Support Vector Machine – AUC using HOG Features\label{fig:SVM-hog img}]{%
		\includegraphics[width=0.32\linewidth]{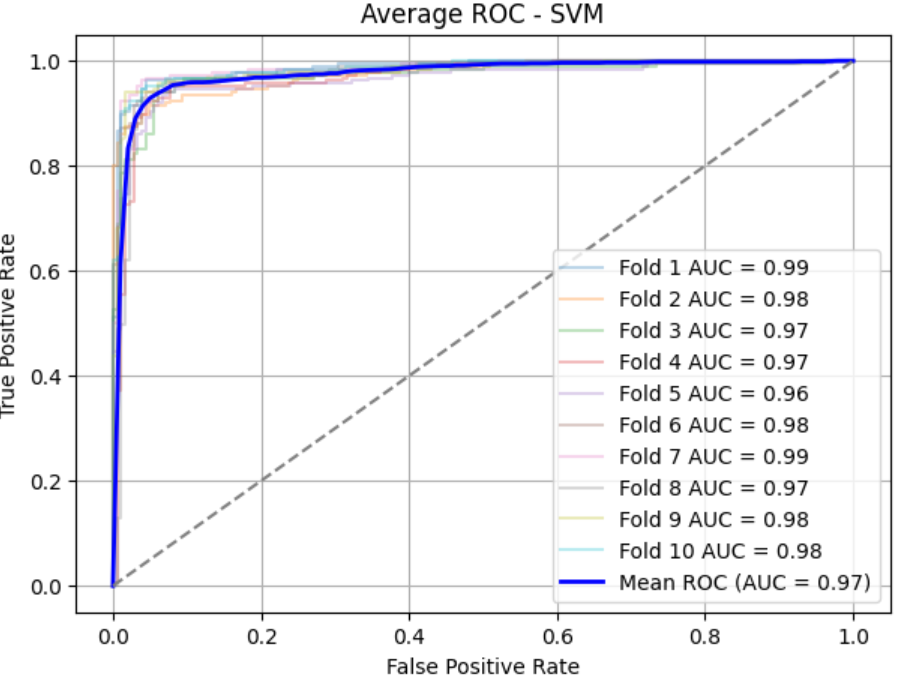}}
	\hfill
	\subfloat[\scriptsize Extra Trees – AUC using HOG Features\label{fig:EXT-hog img}]{%
		\includegraphics[width=0.32\linewidth]{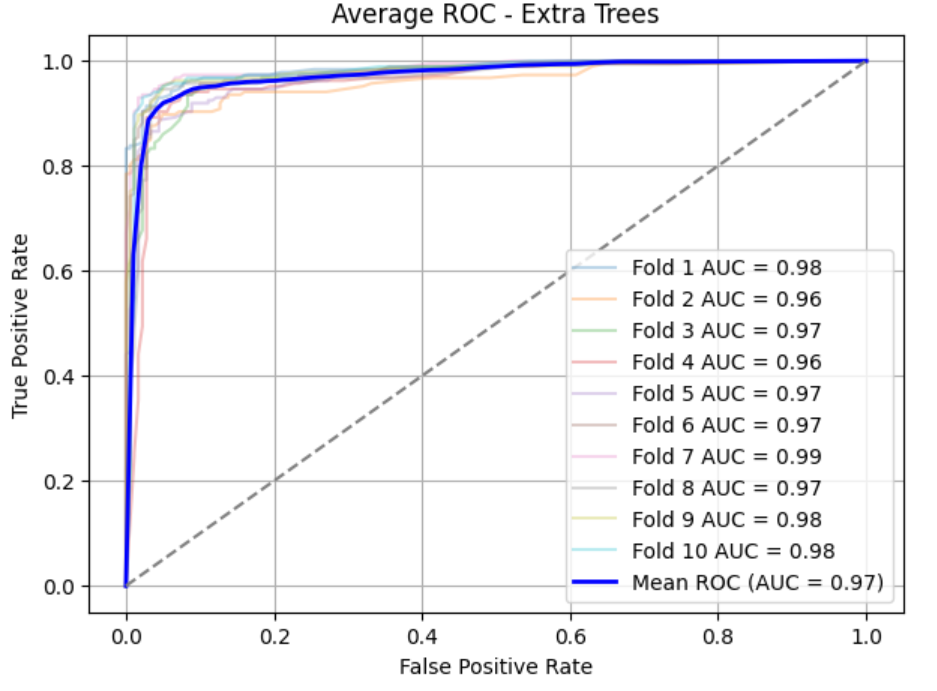}}
	\caption{\footnotesize Comparison of AUC performance for various classifiers using HOG-derived features on the APTOS dataset (binary classification setting).}
	\label{fig:HOG-AUC}
\end{figure*}

\begin{figure*}[t!]
	\centering
	\subfloat[\scriptsize Logistic Regression – Confusion Matrix using HOG Features\label{fig:LR-hog cm}]{%
		\includegraphics[width=0.32\linewidth]{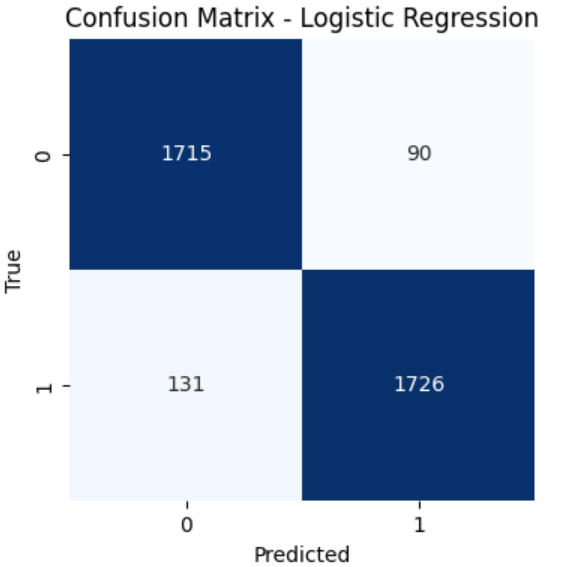}}
	\hfill
	\subfloat[\scriptsize Random Forest – Confusion Matrix using HOG Features\label{fig:RF-hog cm}]{%
		\includegraphics[width=0.32\linewidth]{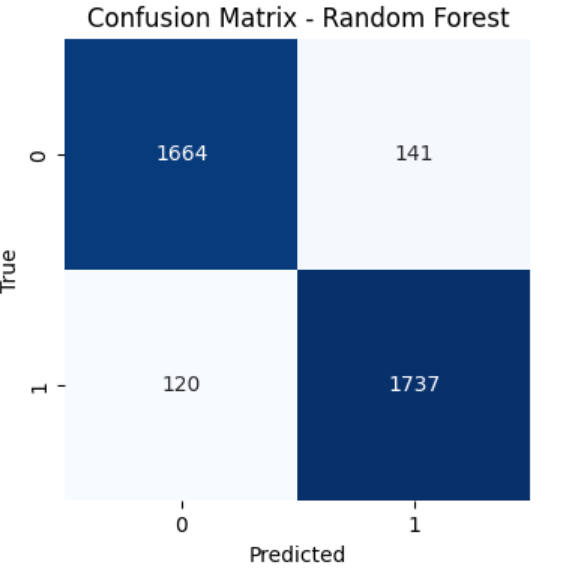}}
	\hfill
	\subfloat[\scriptsize XGBoost – Confusion Matrix using HOG Features\label{fig:XGB-hog cm}]{%
		\includegraphics[width=0.32\linewidth]{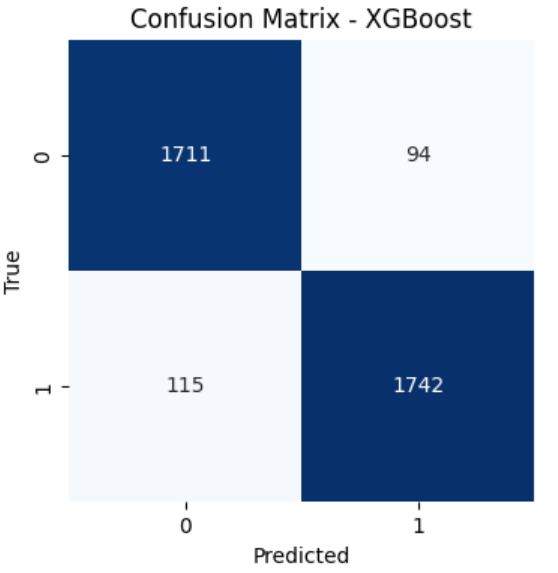}}
	\hfill
	\subfloat[\scriptsize K-Nearest Neighbors – Confusion Matrix using HOG Features\label{fig:KNN-hog cm}]{%
		\includegraphics[width=0.32\linewidth]{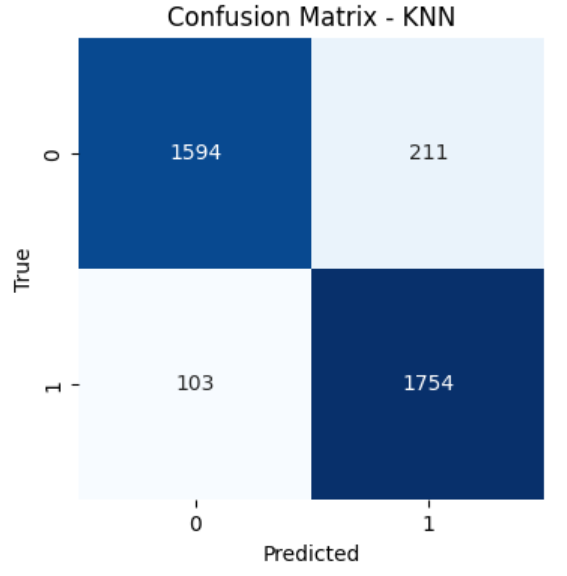}}
	\hfill
	\subfloat[\scriptsize Support Vector Machine – Confusion Matrix using HOG Features\label{fig:SVM-hog cm}]{%
		\includegraphics[width=0.32\linewidth]{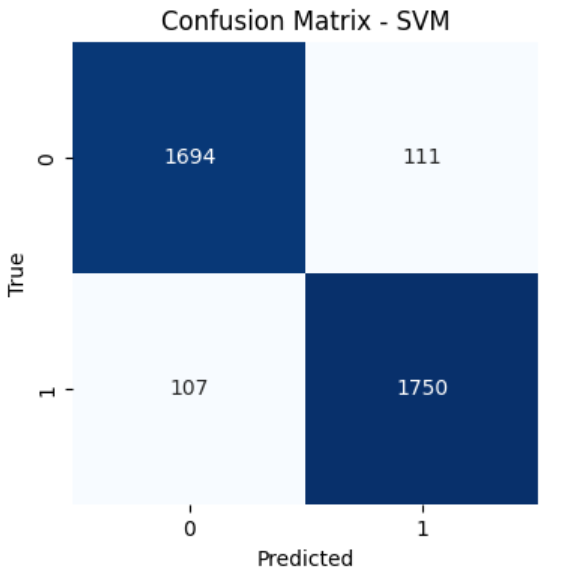}}
	\hfill
	\subfloat[\scriptsize Extra Trees – Confusion Matrix using HOG Features\label{fig:EXT-hog cm}]{%
		\includegraphics[width=0.32\linewidth]{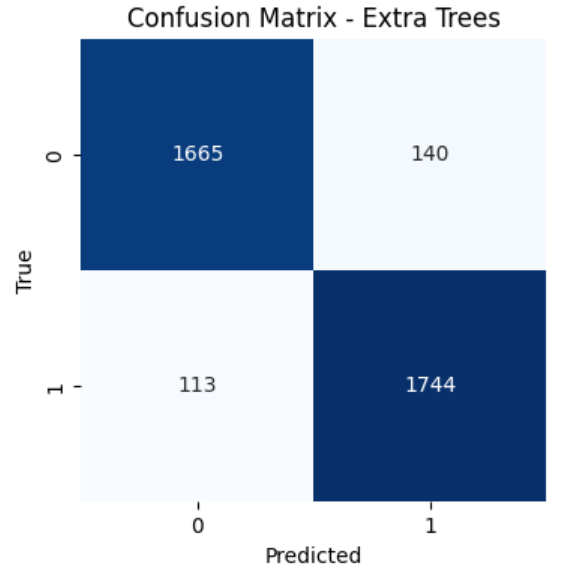}}
	\caption{\footnotesize Confusion matrix comparison of various classifiers trained on HOG-derived features using the APTOS dataset (binary classification setting).}
	\label{fig:HOG-CM}
\end{figure*}

\subsection{Five-Class Classification: DR Severity Levels}

We extend our evaluation to a more challenging five-class classification setting, where the task is to categorize retinal fundus images according to diabetic retinopathy severity levels. Similar to the binary setting, seven traditional machine learning models are trained and tested using the extracted TDA and HOG features. Performance metrics including accuracy, precision, recall, and F1-score are reported as averages with standard deviations over 10-fold cross-validation. Tables~\ref{tab:tda_5class} and~\ref{tab:hog_5class} summarize these results.

\subsubsection*{Comparative Analysis on five-class setting: TDA vs HOG Features}

In the five-class DR severity classification, both TDA and HOG feature sets provide competitive results across different classifiers. HOG features generally lead to higher average accuracies and precision values. For example, Logistic Regression achieves an accuracy of 74.09\% with HOG, significantly surpassing the 63.95\% accuracy achieved with TDA. Similarly, K-Nearest Neighbors (KNN) and Support Vector Machine (SVM) show consistent improvements with HOG features over TDA in terms of recall and F1-score.

However, ensemble models such as XGBoost and Extra Trees yield comparable performances for both feature sets, with accuracies around 74\% in either case, demonstrating that both methods are viable for multi-class classification in retinal images. Decision Tree classifiers perform slightly better with TDA features, but with a noticeable drop in accuracy compared to other models.

Overall, while HOG features offer an edge in linear models and show slightly more stable performance, TDA features remain competitive, especially when paired with ensemble methods. This suggests that both feature extraction techniques capture complementary information useful for distinguishing multiple stages of diabetic retinopathy severity.

\begin{figure*}[t!]
    \centering
    \subfloat[\scriptsize APTOS 2019 Dataset (Binary setting): TDA Features-based Models \label{fig:APTOS(Binary)-TDA}]{%
        \includegraphics[width=0.49\linewidth]{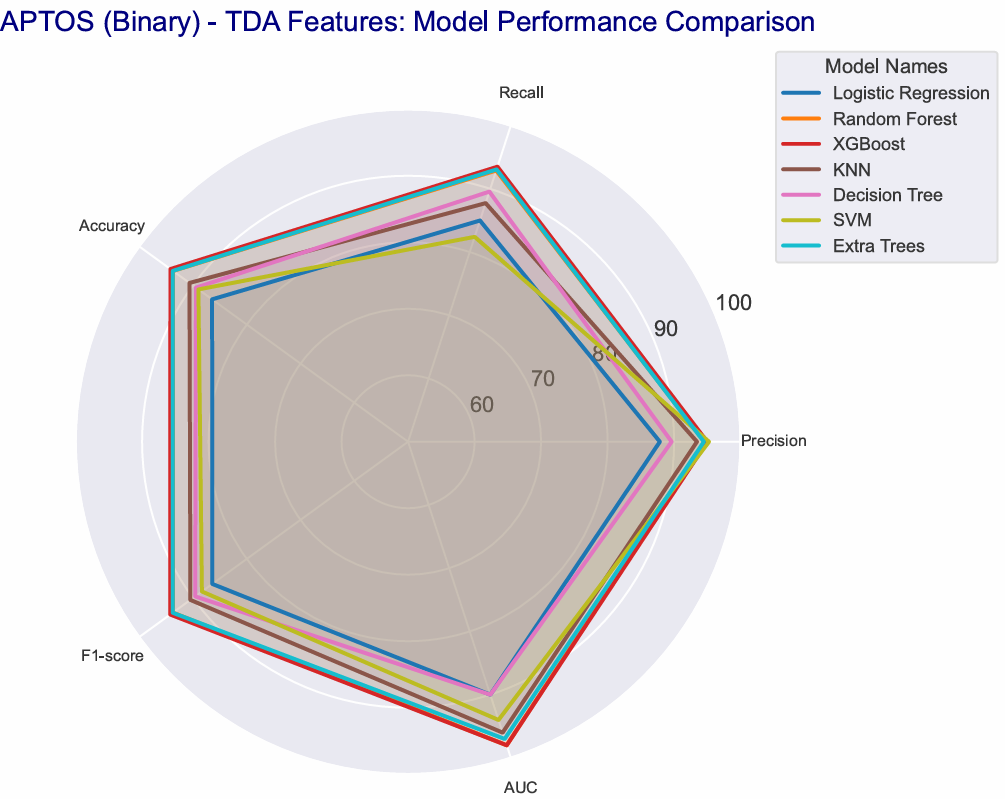}}
    \hfill
    \subfloat[\scriptsize APTOS 2019 Dataset (Binary setting): HOG Features-based Models \label{fig:APTOS(Binary)-HOG}]{%
        \includegraphics[width=0.49\linewidth]{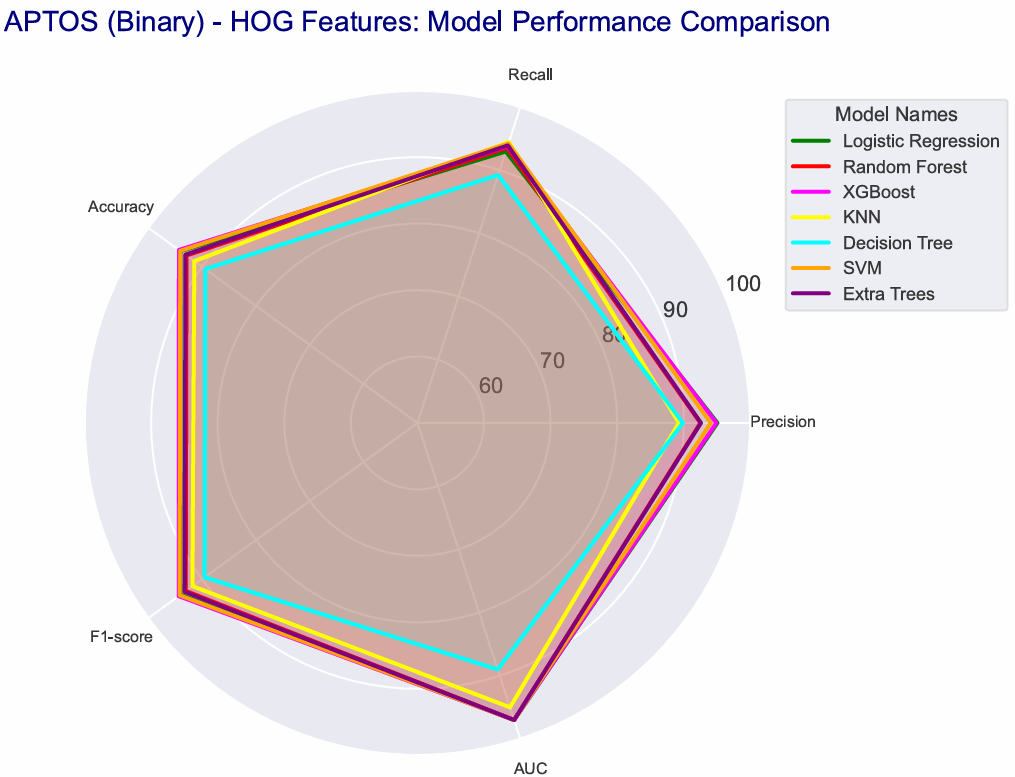}}
    \caption{\footnotesize 
    Radar plots comparing the performance of various classification models on the APTOS 2019 dataset under a binary classification setting using 10-fold cross-validation. 
    (a) Models trained using topological data analysis (TDA) features. 
    (b) Models trained using histogram of oriented gradients (HOG) features. 
    Metrics include Precision, Recall, Accuracy, AUC and F1-score.
    }
    \label{fig:aptos_binary radar}
\end{figure*}

\begin{figure*}[t!]
    \centering
    \subfloat[\scriptsize APTOS 2019 Dataset (5-Class setting): TDA Features-based Models \label{fig:APTOS(5-class)-TDA}]{%
        \includegraphics[width=0.49\linewidth]{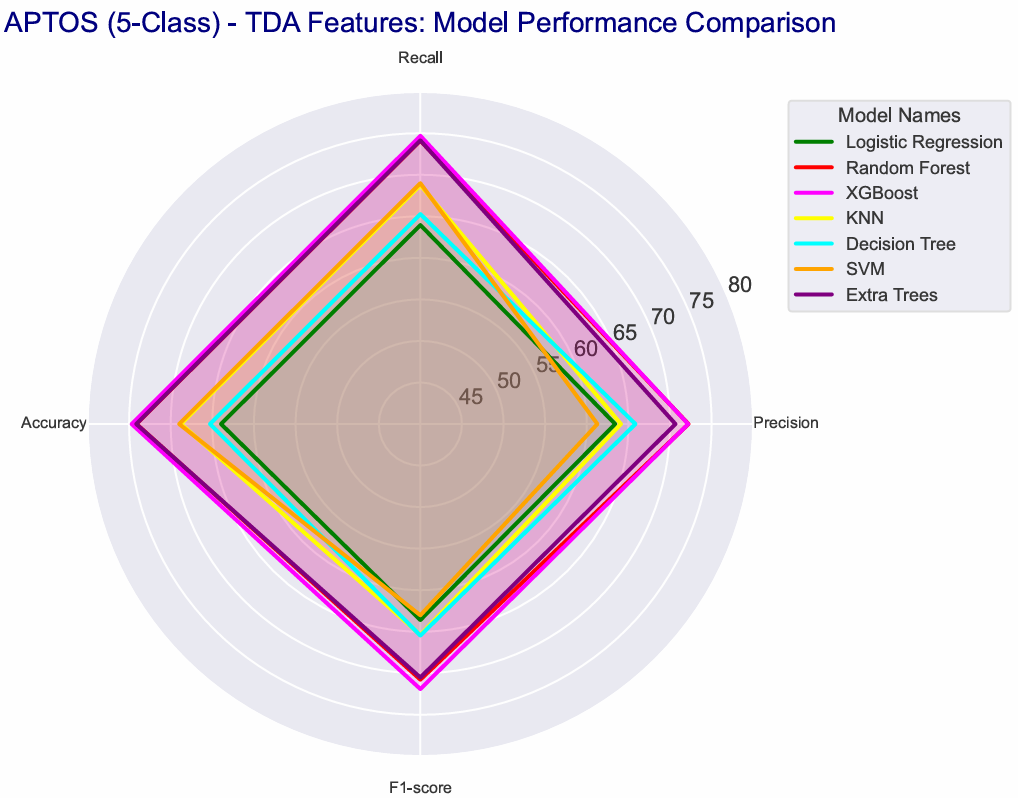}}
    \hfill
    \subfloat[\scriptsize APTOS 2019 Dataset (5-Class setting): HOG Features-based Models \label{fig:APTOS(5-class)-HOG}]{%
        \includegraphics[width=0.49\linewidth]{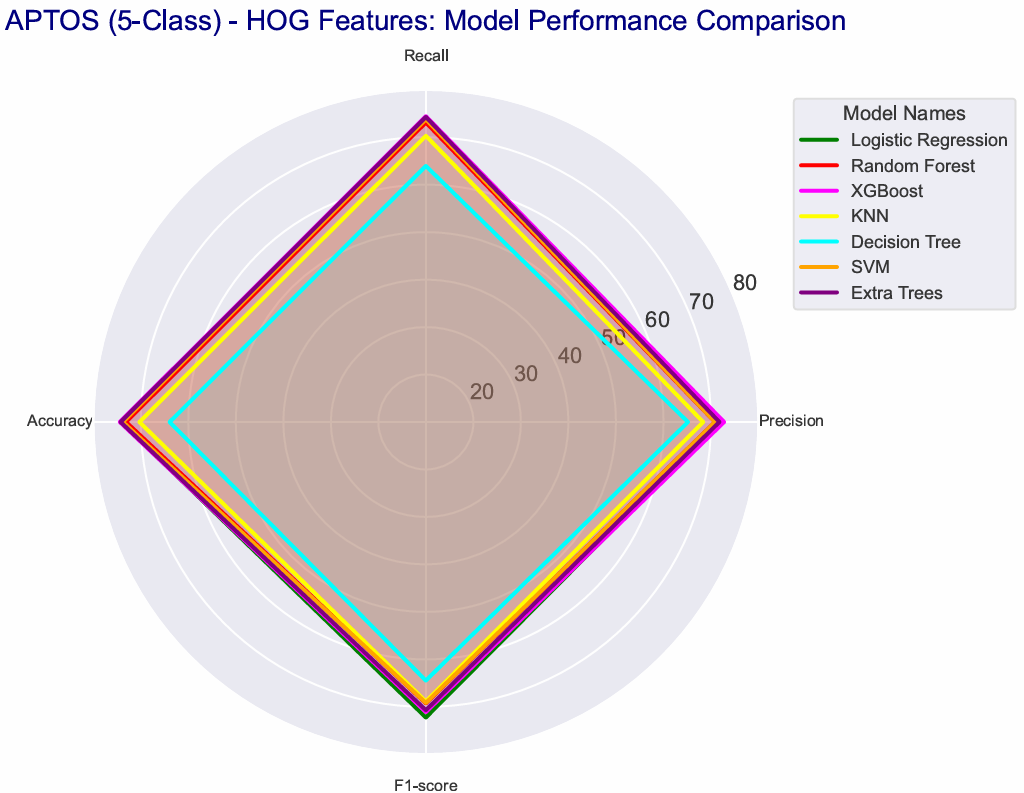}}
    \caption{\footnotesize 
    Radar plots comparing the performance of various classification models on the APTOS 2019 dataset under a 5-Class severity classification setting using 10-fold cross-validation. 
    (a) Models trained using topological data analysis (TDA) features. 
    (b) Models trained using histogram of oriented gradients (HOG) features. 
    Metrics include Precision, Recall, Accuracy, and F1-score.
    }
    \label{fig:aptos_5class radar}
\end{figure*}

\begin{table*}[htbp]
\centering
\caption{APTOS (Binary) - TDA Features}
\label{tab:tda_binary}
\begin{tabular}{lccccc}
\toprule
Model’s Name & Avg Acc & Avg Prec & Avg Rec & Avg F1-score & Avg AUC \\
\midrule
Logistic Regression & 86.43 ± 1.66 & 87.86 ± 1.44 & 84.98 ± 2.30 & 86.39 ± 1.73 & 90.0 \\
Random Forest       & 93.69 ± 0.99 & 94.59 ± 0.96 & 92.89 ± 2.12 & 93.72 ± 1.05 & 98.0 \\
XGBoost             & 94.18 ± 0.92 & 94.98 ± 0.98 & 93.49 ± 1.88 & 94.21 ± 0.96 & 98.0 \\
KNN                 & 90.63 ± 1.03 & 93.44 ± 1.60 & 87.73 ± 1.94 & 90.47 ± 1.07 & 96.0 \\
Decision Tree       & 89.43 ± 0.89 & 89.63 ± 1.34 & 89.55 ± 1.56 & 89.57 ± 0.90 & 90.0 \\
SVM                 & 88.97 ± 1.01 & 95.25 ± 1.54 & 82.40 ± 2.46 & 88.32 ± 1.19 & 94.0 \\
Extra Trees         & 93.72 ± 1.08 & 94.48 ± 0.97 & 93.06 ± 1.85 & 93.75 ± 1.12 & 97.0 \\
\bottomrule
\end{tabular}
\end{table*}

\vspace{1cm}

\begin{table*}[htbp]
\centering
\caption{APTOS (Binary) - HOG Features}
\label{tab:hog_binary}
\begin{tabular}{lccccc}
\toprule
Model’s Name & Avg Acc  & Avg Prec  & Avg Rec  & Avg F1-score  & Avg AUC  \\
\midrule
Logistic Regression & 93.97 ± 1.09 & 95.06 ± 1.30 & 92.95 ± 1.86 & 93.98 ± 1.11 & 97.0 \\
Random Forest       & 92.87 ± 1.37 & 92.51 ± 1.46 & 93.54 ± 1.83 & 93.01 ± 1.35 & 97.0 \\
XGBoost             & 94.29 ± 1.18 & 94.90 ± 1.25 & 93.81 ± 1.95 & 94.34 ± 1.20 & 97.0 \\
KNN                 & 91.43 ± 1.21 & 89.28 ± 1.35 & 94.46 ± 1.84 & 91.78 ± 1.18 & 95.0 \\
Decision Tree       & 89.40 ± 1.94 & 89.88 ± 2.70 & 89.23 ± 2.40 & 89.52 ± 1.89 & 89.0 \\
SVM                 & 94.05 ± 1.06 & 94.07 ± 1.56 & 94.24 ± 2.01 & 94.13 ± 1.06 & 97.0 \\
Extra Trees         & 93.09 ± 1.34 & 92.59 ± 1.32 & 93.92 ± 2.26 & 93.23 ± 1.35 & 97.0 \\
\bottomrule
\end{tabular}
\end{table*}

\begin{table*}[htbp]
\centering
\caption{APTOS (5-Class) - TDA Features}
\label{tab:tda_5class}
\begin{tabular}{lcccc}
\toprule
Model’s Name & Avg Acc  & Avg Prec  & Avg Rec  & Avg F1-score  \\
\midrule
Logistic Regression & 63.95 ± 2.24 & 63.41 ± 2.20 & 63.95 ± 2.24 & 63.59 ± 2.21 \\
Random Forest       & 74.11 ± 1.33 & 72.22 ± 2.53 & 74.11 ± 1.33 & 70.77 ± 1.54 \\
XGBoost             & 74.69 ± 1.52 & 72.18 ± 2.79 & 74.69 ± 1.52 & 71.90 ± 1.84 \\
KNN                 & 68.76 ± 1.44 & 64.10 ± 2.23 & 68.76 ± 1.44 & 65.38 ± 1.53 \\
Decision Tree       & 65.27 ± 2.33 & 65.84 ± 2.13 & 65.27 ± 2.33 & 65.46 ± 2.21 \\
SVM                 & 68.98 ± 1.09 & 61.27 ± 2.97 & 68.98 ± 1.09 & 63.02 ± 1.15 \\
Extra Trees         & 74.14 ± 1.40 & 70.65 ± 2.40 & 74.14 ± 1.40 & 70.51 ± 1.80 \\

\bottomrule
\end{tabular}
\end{table*}

\vspace{1cm}

\begin{table*}[htbp]
\centering
\caption{APTOS (5-Class) - HOG Features}
\label{tab:hog_5class}
\begin{tabular}{lcccc}
\toprule
Model’s Name & Avg Acc  & Avg Prec  & Avg Rec  & Avg F1-score  \\
\midrule
Logistic Regression & 74.09 ± 1.56 & 71.85 ± 2.11 & 74.09 ± 1.56 & 72.24 ± 1.74 \\
Random Forest       & 73.10 ± 1.50 & 70.87 ± 3.12 & 73.10 ± 1.50 & 69.26 ± 1.83 \\
XGBoost             & 74.41 ± 1.02 & 72.76 ± 2.19 & 74.41 ± 1.02 & 71.08 ± 1.10 \\
KNN                 & 70.29 ± 1.24 & 68.35 ± 1.53 & 70.29 ± 1.24 & 68.87 ± 1.29 \\
Decision Tree       & 63.93 ± 2.28 & 65.16 ± 2.22 & 63.93 ± 2.28 & 64.47 ± 2.14 \\

SVM                 & 73.76 ± 1.03 & 70.87 ± 5.16 & 73.76 ± 1.03 & 69.17 ± 1.14 \\
Extra Trees         & 74.25 ± 1.38 & 71.77 ± 2.97 & 74.25 ± 1.38 & 70.69 ± 1.76 \\
\bottomrule
\end{tabular}
\end{table*}

\section{Interpretation of Topological and HOG Features} \label{sec:explain}
\subsection{HOG Features}
Histogram of Oriented Gradients (HOG) features have shown solid performance in our diabetic retinopathy classification experiments, offering a handcrafted representation of local edge and texture information. However, despite their discriminative power, HOG features remain inherently difficult to interpret in a medical imaging context.

This challenge stems from several factors. First, the transformation of each image into a high-dimensional feature vector ($\mathbb{R}^{26{,}244}$) results in an abstract numerical representation where each dimension corresponds to gradient orientations within small image patches. These values cannot be easily mapped back to specific anatomical structures (e.g., optic disc or macula), limiting the ability to understand what visual cues the model relies on for its predictions.

Second, the conversion to grayscale and the local nature of HOG descriptors eliminate color information and broader spatial context—both of which are often critical for clinical diagnosis. As a result, medically meaningful patterns such as lesions, exudates, or hemorrhages may not be explicitly or intuitively captured.
\begin{figure}[H]
    \centering
    \includegraphics[width=\linewidth]{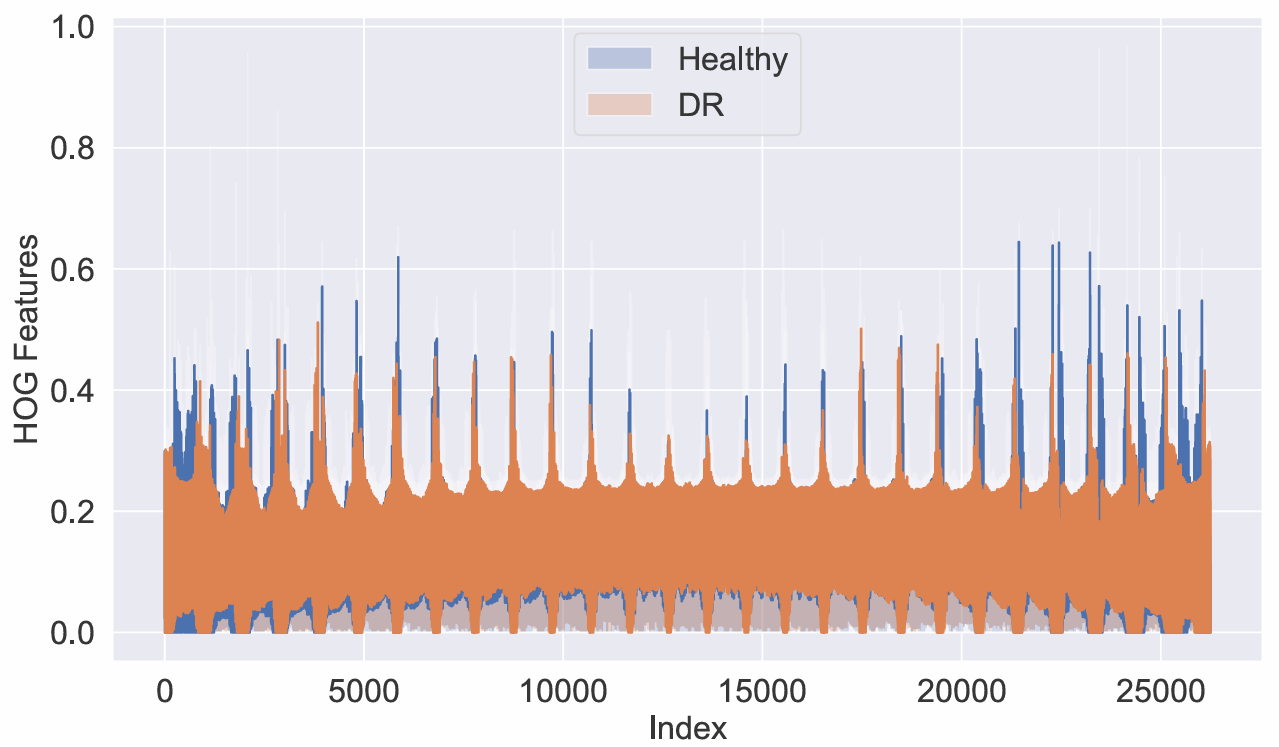}
    \caption{\footnotesize Confidence band visualization (40\%) of HOG feature distributions for diabetic retinopathy (DR) detection using the APTOS 2019 dataset in a binary classification setting. The $x$-axis corresponds to the HOG feature indices, while the $y$-axis represents their normalized values. Overlayed distributions illustrate structural distinctions between Healthy and DR samples. }
    \label{fig:hog-features}
\end{figure}
Lastly, as shown in Figure~\ref{fig:hog-features}, the distribution of HOG features across Healthy and DR classes exhibits substantial overlap. Although some separation exists in localized regions, the overall feature landscape is complex and lacks clearly interpretable distinctions. This reinforces the notion that while HOG encodes useful structural features, it does so in a form that resists intuitive understanding and clinical traceability.

\subsection{Topological Features}
In contrast, Topological Data Analysis (TDA) offers interpretable features rooted in the mathematical properties of shapes and structures within images. As emphasized in the introduction, one of the major strengths of our approach lies in this interpretability, which is critical for trust and transparency in clinical applications. To illustrate the interpretability of topological features, we present in \Cref{fig:aptos_all} the median Betti curves and 40\% non-parametric confidence bands for each class retinal diseases diabetic retinopathy (DR). These visualizations provide a summary of the topological patterns present in normal and abnormal fundus images. The confidence bands and median curves are constructed using techniques outlined in~\cite{gibbons2014nonparametric}.

From a machine learning standpoint, the distinct patterns in \Cref{fig:aptos_all} highlight the discriminative strength of the TDA-derived feature vectors. Each image is transformed into a pair of 100-dimensional Betti-0 and Betti-1 vectors by evaluating topological features across 100 grayscale thresholds uniformly distributed over the interval $[0, 255]$. This results in each image being mapped to a point in the high-dimensional latent space $\mathbb{R}^{100}$. Specifically, for an image $\mathbf{X}$, its topological representation is denoted by $\beta(\mathbf{X}) \in \mathbb{R}^{100}$. The median Betti curves in the figures can thus be interpreted as centroids of class-specific clusters in this latent space, while inter-class differences correspond to the separations between these clusters.

Although the visual differences between Betti curves might appear small, each individual value $\beta_k(t_i)$ in the vector $\beta_k(\mathbf{X}) = [\beta_k(1)\ \beta_k(2)\ \dots \beta_k(100)]$ contributes to the geometric positioning of the point in $\mathbb{R}^{100}$. Consequently, even modest differences between curves translate to significant distances in this high-dimensional space—distances that machine learning models can effectively exploit during training. This separation explains the model's high accuracy and robustness in distinguishing between disease states.

To further ground this interpretability, we provide background on Betti curves—a central tool in our feature extraction pipeline. Unlike other methods of vectorizing persistence diagrams, Betti curves offer a direct and intuitive understanding of the image topology. Betti-0 tracks the number of connected components, and Betti-1 tracks the number of loops or holes, as the pixel intensity threshold increases from black (0) to white (255).

In concrete terms, for any threshold $t_0 \in [0, 255]$, the value $\beta_0(t_0)$ represents the number of disconnected components in the binary image $\X_{t_0}$, while $\beta_1(t_0)$ indicates the number of enclosed loops or holes. These counts—depicted on the vertical axes in our plots—are absolute and not normalized, providing a faithful representation of structural complexity.
{\scriptsize
\begin{figure}[H]
    \centering
    \subfloat[APTOS 2019 Dataset: Betti-0 Curves \label{fig:APTOS-B0}]{%
        \includegraphics[width=0.49\linewidth]{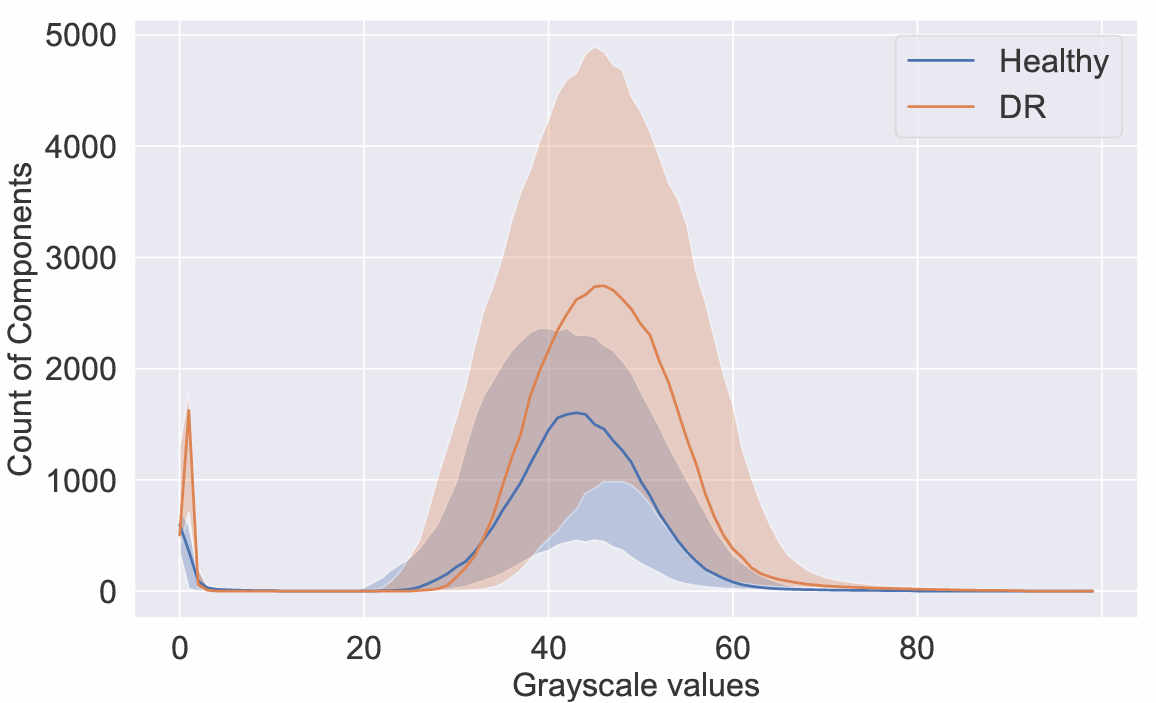}}
    \hfill
    \subfloat[APTOS 2019 Dataset: Betti-1 Curves \label{fig:APTOS-B1}]{%
        \includegraphics[width=0.49\linewidth]{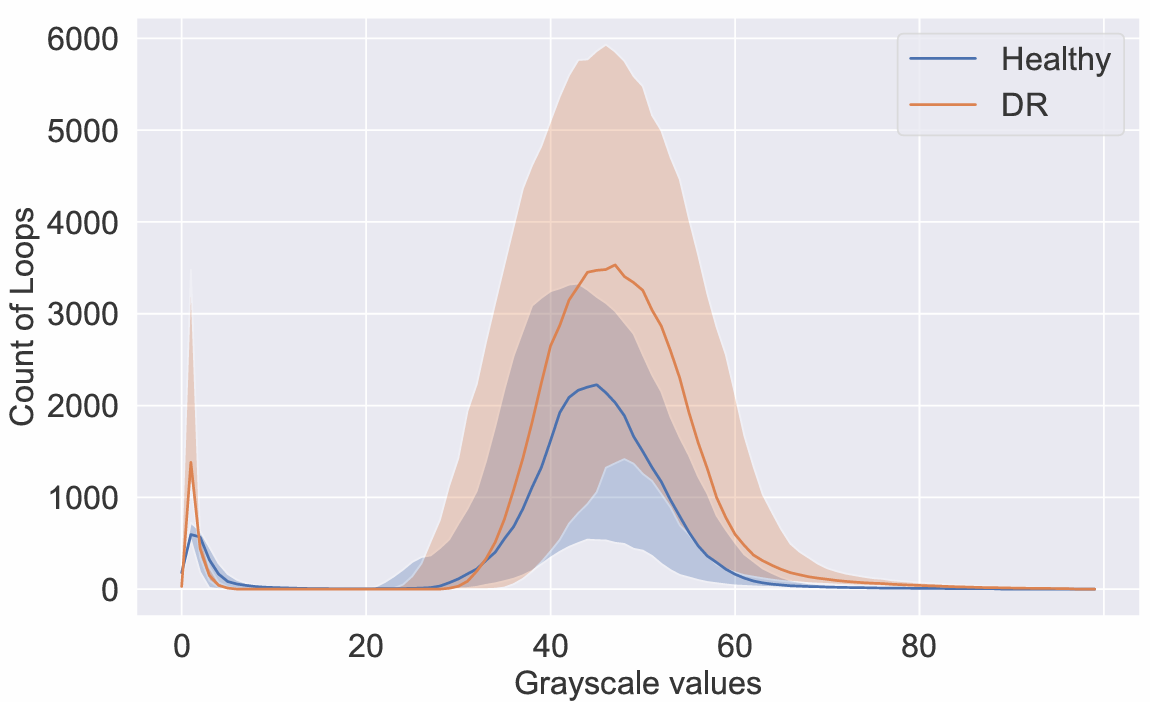}}
    \caption{Median topological feature curves and corresponding $40\%$ confidence bands for diabetic retinopathy (DR) detection using the APTOS 2019 dataset. The $x$-axis denotes color intensity values, while $\beta_0(t)$ and $\beta_1(t)$ represent the number of connected components and the number of one-dimensional loops in the sublevel sets $\mathcal{X}_t$, respectively. \label{fig:aptos_all}}
\end{figure}
}

For instance, in \Cref{fig:APTOS-B1}, we present the Betti-1 curves for the two classes (DR vs. normal) in the APTOS 2019 dataset. We observe that between grayscale values 100 and 150, the loop count in DR images is nearly twice that of normal images. To interpret this, consider two grayscale fundus images, $\X$ (normal) and $\Y$ (DR). At threshold 125, the binary image $\X_{125}$ contains approximately 1,000 loops, whereas $\Y_{125}$ contains about 2,200. This suggests that DR images exhibit a greater number of light-intensity regions (i.e., brighter lesions), indicative of the disease's pathological manifestations. These regions appear as holes in the binarized image and are effectively captured by the Betti-1 curve.

Finally, topological features not only provide competitive performance metrics but also offer a unique window into the structural differences between normal and diseased fundus images. This interpretability is essential for building transparent and clinically trustworthy AI systems.

\section{Discussion} \label{sec:discussion}

This study presents a comprehensive comparative analysis of two distinct feature extraction methods—Topological Data Analysis (TDA) via persistent homology and Histogram of Oriented Gradients (HOG)—applied to retinal fundus images for diabetic retinopathy (DR) classification. Our experiments on the publicly available APTOS dataset, using both binary (Normal vs DR) and multi-class (five-class DR severity) classification settings, reveal several important insights regarding the strengths and limitations of these feature representations.

In the binary classification setting, HOG features consistently outperform TDA features across most traditional machine learning models, especially linear classifiers such as Logistic Regression and Support Vector Machines. The superior discriminative power of HOG is reflected in higher average accuracy, precision, recall, and F1-scores. This suggests that gradient-based local texture patterns captured by HOG are highly effective in distinguishing normal from diseased retinal images. Meanwhile, tree-based ensemble models like XGBoost and Random Forest show competitive performance with both feature sets, indicating their robustness to different feature representations. Notably, TDA features occasionally achieve slightly better AUC scores, hinting at their potential for capturing subtle topological patterns that aid in class separability.

For the more challenging five-class classification of DR severity levels, both TDA and HOG features maintain competitive performance, though HOG features generally maintain a performance edge in linear models. Ensemble methods continue to provide stable and relatively high accuracies with both feature types, demonstrating their adaptability to complex multi-class tasks. Interestingly, the gap between TDA and HOG narrows in this setting, emphasizing that TDA’s topological features capture meaningful structural information relevant for distinguishing different disease stages, despite being fewer in number compared to the high-dimensional HOG descriptors.

From a practical perspective, these findings highlight complementary strengths: HOG features excel at capturing local texture and gradient information, which is crucial for detecting visible retinal abnormalities, while TDA features offer a more abstract, shape-based characterization of retinal structures that can capture global organizational patterns. Additionally, the lower dimensionality of TDA features can offer advantages in terms of computational efficiency and interpretability, which are valuable in clinical applications where transparency and speed are critical.

Overall, this study demonstrates that both TDA and HOG provide viable feature extraction approaches for retinal disease classification and can be integrated with various machine learning pipelines. Future work will explore hybrid architectures that combine these complementary features with deep learning models to further enhance classification accuracy and robustness, as well as extending the analysis to other medical diseases and imaging modalities.

\section{Limitations}

While this study offers valuable insights into the comparative performance of TDA and HOG features for retinal disease classification, several limitations should be acknowledged. The analysis focuses on traditional machine learning models and does not fully explore end-to-end deep learning frameworks, which may be better suited to capturing complex feature interactions. Although hybrid approaches were discussed, actual integration of TDA and HOG features into deep learning pipelines remains a direction for future work.

The dataset used (APTOS), though publicly available and relatively large, may not encompass the full spectrum of retinal pathologies across diverse populations and imaging conditions, which could affect the generalizability of the findings.

In terms of methodology, the TDA features were limited to persistent homology on cubical complexes; other topological descriptors or filtration strategies might capture additional meaningful patterns. Likewise, HOG features, while effective, are handcrafted and may overlook subtle image characteristics that learned features could potentially capture.

Finally, this work did not extensively evaluate the computational efficiency and scalability of TDA feature extraction, particularly for high-resolution images or large-scale datasets. Future studies should address these aspects to better assess the feasibility of deploying such methods in real-world clinical environments.

\section{Conclusion} \label{sec:conclusion}

In this work, we first conducted a comprehensive comparative study of two feature extraction techniques—Topological Data Analysis (TDA) using persistent homology and Histogram of Oriented Gradients (HOG)—for the classification of retinal fundus images in diabetic retinopathy diagnosis. Using the large, well-known, publicly available APTOS dataset, we evaluated these features under both binary (Normal vs DR) and five-class DR severity classification settings. We extracted 800 TDA features and 26,244 HOG features per image and employed seven traditional machine learning models to assess classification performance.

Our results demonstrate that HOG features generally outperform TDA features in linear and kernel-based classifiers, achieving higher accuracy, precision, recall, and F1-scores, particularly in the binary classification setting. However, TDA features remain competitive, especially when combined with ensemble methods, and offer complementary strengths in capturing the topological structure of retinal images. Both feature types showed robust performance in the more challenging multi-class setting, highlighting their utility for detailed disease severity assessment.

Importantly, both TDA and HOG features can be seamlessly integrated into deep learning frameworks to further enhance model performance, suggesting promising avenues for hybrid approaches. This study underscores the potential of combining traditional handcrafted and topological features to improve the accuracy and interpretability of retinal disease classification systems.

\section{Future Work}

Building on the current findings, future research will focus on extending the generalizability of this framework to a wider range of medical imaging modalities and disease contexts beyond retinal analysis. A key direction involves integrating the gradient-based and topological feature extraction techniques within advanced deep learning architectures to capture more complex and hierarchical image representations. Additionally, efforts will be made to evaluate the approach on larger, more diverse, and multi-center clinical datasets to enhance robustness and generalizability across populations and imaging devices. Addressing computational efficiency and scalability of the feature extraction processes, especially for high-resolution images and extensive datasets, will also be prioritized to facilitate practical deployment in real-world clinical workflows. These advancements are critical to establishing the framework’s clinical relevance and enabling its adoption in routine medical image analysis.

 \section{Author's Contribution} \label{sec:authors_contribution}

 Faisal Ahmed conceived the study and was responsible for data acquisition, coding, and data analysis. He drafted, reviewed, and revised the manuscript, and approved the final version for submission.

 \section{Funding} \label{sec:authors_funding}
 The author received no financial support for the research, authorship, or publication of this work.

\clearpage

\bibliographystyle{elsarticle-num-names}

\bibliography{refs}

@article{ahmed2025topo,
  title={Topo-CNN: Retinal Image Analysis with Topological Deep Learning},
  author={Ahmed, Faisal and Bhuiyan, Mohammad Alfrad Nobel and Coskunuzer, Baris},
  journal={Journal of Imaging Informatics in Medicine},
  pages={1--17},
  year={2025},
  publisher={Springer}
}

@article{litjens2017survey,
  title={A survey on deep learning in medical image analysis},
  author={Litjens, Geert and Kooi, Thijs and Bejnordi, Babak Ehteshami and Setio, Arnaud Arindra Adiyoso and Ciompi, Francesco and Ghafoorian, Mohsen and van der Laak, Jeroen AWM and van Ginneken, Bram and Sánchez, Clara I},
  journal={Medical image analysis},
  volume={42},
  pages={60--88},
  year={2017},
  publisher={Elsevier}
}

@article{erickson2017machine,
  title={Machine learning for medical imaging},
  author={Erickson, Bradley J and Korfiatis, Petros and Akkus, Zeynettin and Kline, Timothy L},
  journal={Radiographics},
  volume={37},
  number={2},
  pages={505--515},
  year={2017},
  publisher={Radiological Society of North America}
}

@article{lundervold2019overview,
  title={An overview of deep learning in medical imaging focusing on MRI},
  author={Lundervold, Alexander Selvikvåg and Lundervold, Arvid},
  journal={Zeitschrift für Medizinische Physik},
  volume={29},
  number={2},
  pages={102--127},
  year={2019},
  publisher={Elsevier}
}

@inproceedings{reininghaus2015stable,
  title={A stable multi-scale kernel for topological machine learning},
  author={Reininghaus, Jan and Huber, Stefan and Bauer, Ulrich and Kwitt, Roland},
  booktitle={Proceedings of the IEEE conference on computer vision and pattern recognition},
  pages={4741--4748},
  year={2015}
}

@article{pun2022topological,
  title={Topological data analysis for medical imaging: A review of the state-of-the-art and a new perspective},
  author={Pun, Chi Seng and Chung, Albert C S},
  journal={Medical Image Analysis},
  volume={77},
  pages={102355},
  year={2022},
  publisher={Elsevier}
}

@inproceedings{dalal2005histograms,
  title={Histograms of oriented gradients for human detection},
  author={Dalal, Navneet and Triggs, Bill},
  booktitle={CVPR},
  pages={886--893},
  year={2005},
  organization={IEEE}
}

@article{ashraf2021hog,
  title={HOG feature-based classification of retinal diseases using fundus images},
  author={Ashraf, Ahsan and others},
  journal={Biomedical Signal Processing and Control},
  volume={68},
  pages={102741},
  year={2021}
}

@article{al2022retinal,
  title={Retinal disease detection using HOG and machine learning classifiers},
  author={Al-Bashir, Ahmad and others},
  journal={Computers in Biology and Medicine},
  volume={146},
  pages={105693},
  year={2022}
}

@article{zhang2017hybrid,
  title={A hybrid method for cervical cancer diagnosis using HOG and SVM},
  author={Zhang, Y. and others},
  journal={Computers in Biology and Medicine},
  volume={85},
  pages={134--144},
  year={2017}
}

@article{irshad2013methods,
  title={Methods for evaluating histopathological image segmentation: breast cancer segmentation and classification},
  author={Irshad, Humayun},
  journal={Medical Image Analysis},
  volume={17},
  number={3},
  pages={374--385},
  year={2013}
}

@article{setio2016pulmonary,
  title={Pulmonary nodule detection in CT images: false positive reduction using multi-view convolutional networks},
  author={Setio, Arnaud A A and others},
  journal={IEEE Transactions on Medical Imaging},
  volume={35},
  number={5},
  pages={1160--1169},
  year={2016}
}

@article{adegun2020hybrid,
  title={Hybrid deep learning and HOG feature approach for brain tumor detection in MRI images},
  author={Adegun, A. A. and others},
  journal={Neural Computing and Applications},
  volume={32},
  pages={10983--10995},
  year={2020}
}

@article{kumar2017dataset,
  title={A dataset and a technique for generalised nuclear segmentation for computational pathology},
  author={Kumar, Nishant and others},
  journal={IEEE Transactions on Medical Imaging},
  volume={36},
  number={7},
  pages={1550--1560},
  year={2017}
}

@article{tauzin2021giotto,
  title={giotto-tda:: A topological data analysis toolkit for machine learning and data exploration},
  author={Tauzin, Guillaume and Lupo, Umberto and Tunstall, Lewis and P{\'e}rez, Julian Burella and Caorsi, Matteo and Medina-Mardones, Anibal M and Dassatti, Alberto and Hess, Kathryn},
  journal={Journal of Machine Learning Research},
  volume={22},
  number={39},
  pages={1--6},
  year={2021}
}

@phdthesis{ahmed2023topological,
  title={Topological Machine Learning in Medical Image Analysis},
  author={Ahmed, Faisal},
  year={2023},
  school = {The University of Texas at Dallas},
}

@inproceedings{ahmed2023tofi,
  title={ToFi-ML: Retinal Image Screening with Topological Machine Learning},
  author={Ahmed, Faisal and Coskunuzer, Baris},
  booktitle={Annual Conference on Medical Image Understanding and Analysis},
  pages={281--297},
  year={2023},
  organization={Springer}
}

@INPROCEEDINGS{10385822,
  author={Yadav, Ankur and Ahmed, Faisal and Daescu, Ovidiu and Gedik, Reyhan and Coskunuzer, Baris},
  booktitle={2023 IEEE International Conference on Bioinformatics and Biomedicine (BIBM)}, 
  title={Histopathological Cancer Detection with Topological Signatures}, 
  year={2023},
  volume={},
  number={},
  pages={1610-1619},
  doi={10.1109/BIBM58861.2023.10385822}}

@inproceedings{ahmed2023topo,
  title={Topo-{CXR}: Chest {X}-ray {TB} and {P}neumonia {S}creening with {T}opological {M}achine {L}earning},
  author={Ahmed, Faisal and Nuwagira, Brighton and Torlak, Furkan and Coskunuzer, Baris},
  booktitle={Proceedings of the IEEE/CVF International Conference on Computer Vision},
  pages={2326--2336},
  year={2023}
}

@book{gibbons2014nonparametric,
  title={Nonparametric statistical inference},
  author={Gibbons, Jean Dickinson and Chakraborti, Subhabrata},
  year={2014},
  publisher={CRC press}
}

@article{otter2017roadmap,
  title={A roadmap for the computation of persistent homology},
  author={Otter, Nina and Porter, Mason A and Tillmann, Ulrike and Grindrod, Peter and Harrington, Heather A},
  journal={EPJ Data Science},
  volume={6},
  pages={1--38},
  year={2017},
  publisher={Springer}
}

@misc{tauzin2020giottotda,
      title={giotto-{TDA}: A TDA Toolkit for Machine Learning and Data Exploration},
      author={Guillaume Tauzin and others},
      year={2020},
      eprint={2004.02551},
      archivePrefix={arXiv},
      primaryClass={cs.LG}
}

@inproceedings{milosavljevic2011zigzag,
  title={Zigzag persistent homology in matrix multiplication time},
  author={Milosavljevi{\'c}, Nikola and Morozov, Dmitriy and Skraba, Primoz},
  booktitle={{SoCG}},
  pages={216--225},
  year={2011}
}

@article{camara2016inference,
  title={Inference of ancestral recombination graphs through topological data analysis},
  author={C{\'a}mara, Pablo G and Levine, Arnold J and Rabadan, Raul},
  journal={PLoS computational biology},
  volume={12},
  number={8},
  pages={e1005071},
  year={2016},
  publisher={Public Library of Science San Francisco, CA USA}
}

@article{stolz2021topological,
  title={Topological data analysis of task-based fMRI data from experiments on schizophrenia},
  author={Stolz, Bernadette J and others},
  journal={Journal of Physics: Complexity},
  volume={2},
  number={3},
  pages={035006},
  year={2021},
  publisher={IOP Publishing}
}

@article{qaiser2019fast,
  title={Fast and accurate tumor segmentation of histology images using persistent homology and deep convolutional features},
  author={Qaiser, Talha and Tsang, Yee-Wah and Taniyama, Daiki and Sakamoto, Naoya and Nakane, Kazuaki and Epstein, David and Rajpoot, Nasir},
  journal={Medical image analysis},
  volume={55},
  pages={1--14},
  year={2019},
  publisher={Elsevier}
}

@article{bendich2016persistent,
  title={Persistent homology analysis of brain artery trees},
  author={Bendich, Paul and Marron, James S and Miller, Ezra and Pieloch, Alex and Skwerer, Sean},
  journal={The annals of applied statistics},
  volume={10},
  number={1},
  pages={198},
  year={2016},
  publisher={NIH Public Access}
}

@article{kanari2018topological,
  title={A topological representation of branching neuronal morphologies},
  author={Kanari, Lida and others},
  journal={Neuroinformatics},
  volume={16},
  number={1},
  pages={3--13},
  year={2018},
  publisher={Springer}
}

@article{berry2020functional,
  title={Functional summaries of persistence diagrams},
  author={Berry, Eric and Chen, Yen-Chi and Cisewski-Kehe, Jessi and Fasy, Brittany Terese},
  journal={Journal of Applied and Computational Topology},
  volume={4},
  number={2},
  pages={211--262},
  year={2020},
  publisher={Springer}
}

@article{crawford2020predicting,
  title={Predicting clinical outcomes in glioblastoma: an application of topological and functional data analysis},
  author={Crawford, Lorin and et. al.},
  journal={J.Amer.Stat.Assoc.},
  volume={115},
  number={531},
  pages={1139--1150},
  year={2020},
  publisher={Taylor \& Francis}
}

@article{rieck2020uncovering,
  title={Uncovering the topology of time-varying fMRI data using cubical persistence},
  author={Rieck, Bastian and others},
  journal={{NeurIPS}},
  volume={33},
  pages={6900--6912},
  year={2020}
}

@book{dey2022computational,
  title={Computational Topology for Data Analysis},
  author={Dey, Tamal Krishna and Wang, Yusu},
  year={2022},
  publisher={Cambridge University Press}
}

@book{hatcher2002algebraic,
  title={Algebraic Topology},
  author={Hatcher, Allen},
  year={2002},
  publisher={Cambridge University Press}
}

@article{chazal2021introduction,
  title={An introduction to topological data analysis: fundamental and practical aspects for data scientists},
  author={Chazal, F. and Michel, B.},
  journal={Frontiers in Artificial Intelligence},
  volume={4},
  year={2021}
}

@book{carlsson2021topological,
  title={TDA},
  author={Carlsson, Gunnar and Vejdemo-Johansson, Mikael},
  year={2021},
  publisher={Cambridge University Press}
}

@article{lawson2019persistent,
  title={Persistent homology for the quantitative evaluation of architectural features in prostate cancer histology},
  author={Lawson, Peter and Sholl, Andrew B and Brown, J Quincy and Fasy, Brittany Terese and Wenk, Carola},
  journal={Scientific reports},
  volume={9},
  number={1},
  pages={1139},
  year={2019},
  publisher={Nature Publishing Group UK London}
}

@misc{aptos2019,
  title  = {Asia {P}acific  {T}ele-{O}phthalmology {S}ociety ({APTOS}) 2019 {B}lindness {D}etection {D}ataset},
  author = {APTOS},
  year   = {2019},
  note   = {\url{https://www.kaggle.com/c/aptos2019-blindness-detection}}
}

@article{wasserman2018topological,
  title={TDA},
  author={Wasserman, Larry},
  journal={Annual Review of Statistics and Its Application},
  volume={5},
  pages={501--532},
  year={2018},
  publisher={Annual Reviews}
}

\end{document}